\documentclass[runningheads]{llncs}
\usepackage{graphicx}

\usepackage{tikz}
\usepackage{comment}
\usepackage{amsmath,amssymb} % define this before the line numbering.
\usepackage{color}
\usepackage[width=122mm,left=12mm,paperwidth=146mm,height=193mm,top=12mm,paperheight=217mm]{geometry}

\usepackage{algorithm}
\usepackage{algorithmic}
\usepackage{booktabs}
\usepackage{multirow}
\usepackage{array}
\usepackage{booktabs}
\usepackage{makecell}
\usepackage{float}

\usepackage{diagbox}
\usepackage{cuted} % strip
\makeatletter 
  \newcommand\figcaption{\def\@captype{figure}\caption} 
  \newcommand\tabcaption{\def\@captype{table}\caption} 
\makeatother

\begin{document}

\pagestyle{headings}
\mainmatter
\title{One-stage Video Instance Segmentation: \\ From Frame-in Frame-out to Clip-in Clip-out} % Replace with your title
\titlerunning{CiCo}
\author{Minghan Li \and Zhang Lei}
\authorrunning{Minghan Li and Zhang Lei}
\institute{
             Dept. of Computing, The Hong Kong Polytechnic University\\
	     \email{liminghan0330@gmail.com, cslzhang@comp.polyu.edu.hk}
}

\maketitle
\begin{abstract}
\vspace{-2mm}
Many video instance segmentation (VIS) methods partition a video sequence into individual frames to detect and segment objects frame by frame. However, such a frame-in frame-out (FiFo) pipeline is ineffective to exploit the temporal information. Based on the fact that adjacent frames in a short clip are highly coherent in content, we propose to extend the one-stage FiFo framework to a clip-in clip-out (CiCo) one, which performs VIS clip by clip. Specifically, we stack FPN features of all frames in a short video clip to build a spatio-temporal feature cube, and replace the 2D conv layers in the prediction heads and the mask branch with 3D conv layers, forming clip-level prediction heads (CPH) and clip-level mask heads (CMH). Then the clip-level masks of an instance can be generated by feeding its box-level predictions from CPH and clip-level features from CMH into a small fully convolutional network. A clip-level segmentation loss is proposed to ensure that the generated instance masks are temporally coherent in the clip. The proposed CiCo strategy is free of inter-frame alignment, and can be easily embedded into existing FiFo based VIS approaches. To validate the generality and effectiveness of our CiCo strategy, we apply it to two representative FiFo methods, Yolact \cite{bolya2019yolact} and CondInst \cite{tian2020conditional}, resulting in two new one-stage VIS models, namely CiCo-Yolact and CiCo-CondInst, which achieve 37.1/37.3\%, 35.2/35.4\% and 17.2/18.0\% mask AP using the ResNet50 backbone, and 41.8/41.4\%, 38.0/38.9\% and 18.0/18.2\% mask AP using the Swin Transformer tiny backbone on YouTube-VIS 2019, 2021 and OVIS valid sets, respectively, recording new state-of-the-arts. Code and video demos of CiCo can be found at \url{https://github.com/MinghanLi/CiCo}.

\end{abstract}
\section{Introduction}\label{sec:intro}
\vspace{-1mm}

\begin{figure*}[!t]
\centering
\includegraphics[width=0.99\linewidth]{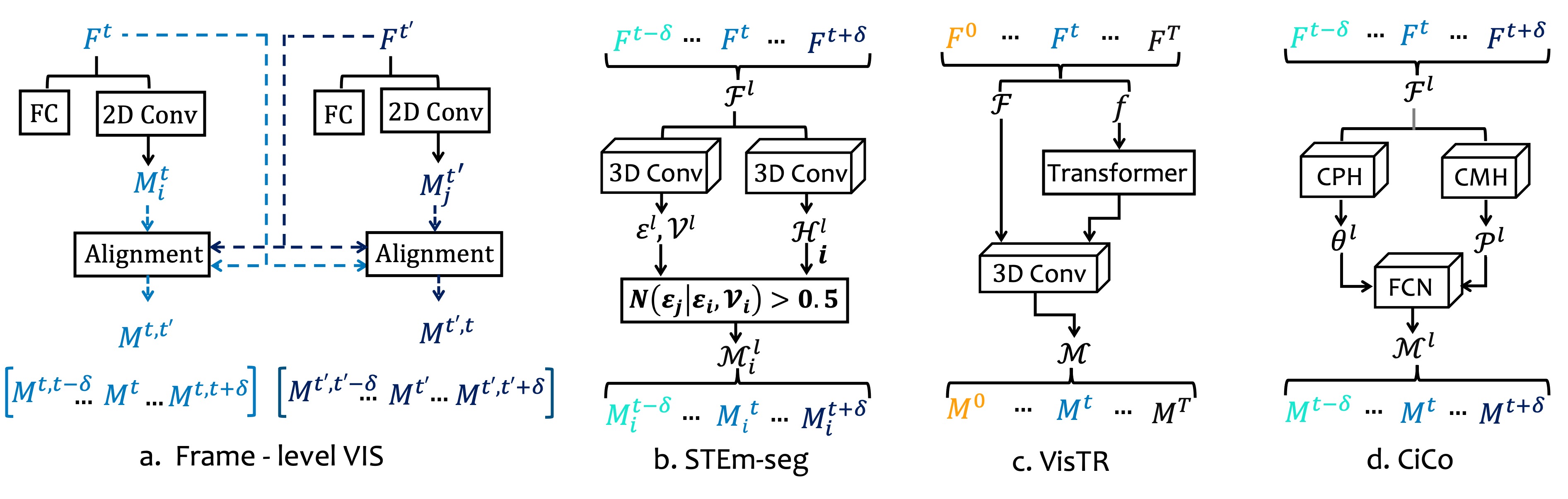}
\vspace{-8mm}
\caption{
Comparison of existing VIS pipelines and our CiCo approach. $f, F, \mathcal{F}$ denote FPN features of shapes $Thw\!\times\!C$, $h\!\times\! w\times\! C$ and $T\! \times\! h\!\times\! w\times C$, respectively, where $T, C$ are the numbers of frames and channels, and $h, w$ are the height and width of features. $M^t$ and $M^{t,t'}$ denote the predicted instance mask on the $t$-th frame and the aligned instance mask from the $t$-th frame to the $t'$-th frame, while $\mathcal{M}$ and $\mathcal{M}^l$ are the predicted clip-level instance masks in the whole video and the $l$-th clip, respectively. In sub-figure (b), $\varepsilon^l$ and $\mathcal{V}^l$ are the pixel-level embedding and variance in a video clip, and $N(\varepsilon_i, \mathcal{V}_i)$ denotes a Gaussian distribution with mean $\varepsilon_i$ and variance $\mathcal{V}_i$. In sub-figure (d), $\theta^l$ represents the box-level parameters of dynamic filters (e.g., weights and bias) in FCN, and $\mathcal{P}^l$ is the clip-level mask features of shape $(2\delta\!+\! 1)\times h\!\times\! w\!\times\! k$. }\label{fig:CiCo}
\vspace{-3mm}
\end{figure*}
Video instance segmentation (VIS)~\cite{yang2019video} aims to obtain pixel-level segmentation masks for individual instances of different classes over the entire video, which needs to simultaneously classify, detect, segment and track objects in a video. Many of the current VIS methods are frame-in frame-out (FiFo) approaches, which first partition the whole video into individual frames to segment objects frame by frame, and then associate the predicted instance masks across frames by integrating temporal information. For example, based on the typical two-stage image instance segmentation (IIS) method Mask R-CNN~\cite{he2017mask}, MaskTrack R-CNN~\cite{yang2019video} adds a new branch to predict embedding vectors of objects and then connect instances across frames. Fig.~\ref{fig:CiCo}(a) illustrates the general pipeline of frame-level VIS methods. However, such a pipeline with inter-frame alignment is inefficient and ineffective to explore the temporal information for more robust classification, detection and segmentation. 

Several approaches have been recently proposed to alleviate the aforementioned issues. As shown in Fig. \ref{fig:CiCo}(b), STEm-seg \cite{Athar_Mahadevan20ECCV} performs spatio-temporal embedding with a bottom-up paradigm, which pulls the pixels belonging to the same object close to each other and pushes pixels of different objects far from each other. Unfortunately, the bottom-up paradigm results in much lower performance than the top-down ones. The VisTR \cite{wang2020end} method, as shown in Fig. \ref{fig:CiCo}(c), reshapes the FPN features of all frames in the video into a sequence and feeds it to a Transformer \cite{vaswani2017attention} to generate global spatio-temporal attention for each object, thereby achieving instance segmentation of the entire video. To reduce the computational and storage overhead of VisTR, IFC \cite{hwang2021video} has been recently developed, which employs separately spatial and temporal transformers to learn attention on a clip, achieving state-of-the-art accuracy.

Actually, the contents of adjacent frames in a short time period are highly coherent, which can be better exploited to detect and segment objects in a video. In Fig. \ref{fig:TIoU}, we plot some statistics on the temporal intersection over union of bounding boxes and masks by using the YouTube-VIS 2019 train set. It can be observed that the locations of many instances in one frame only slightly deviate from that in adjacent frames. In addition, the multiple frames in a video could provide more comprehensive viewpoints on the appearance of an object, and hence provide richer spatio-temporal contextual information for object tracking.

By taking advantage of the temporal coherence in a video, we propose a general strategy to extend the frame-in frame-out (FiFo) one-stage VIS pipeline to a clip-in clip-out (CiCo) one, which is able to generate instance masks clip by clip. Specifically, as shown in Fig. \ref{fig:CiCo}(d), we stack all FPN features of all frames in a short video clip to build a spatio-temporalfeature cube, and replaces the 2D convolutional ({\it{conv}}) layers in the prediction heads and the mask head with 3D {\it{conv}} layers, termed clip-level prediction heads (CPH) and clip-level mask heads (CMH). CPH learns the box-level predictions, including class confidences, regression coordinates of bounding boxes, or mask parameters shared by an object throughout the clip, while CMH generates a set of clip-level feature maps, also called prototypes in Yolact \cite{bolya2019yolact}. Then, the final clip-level masks of an object through the clip can be obtained by feeding the clip-level prototypes and its corresponding box-level predictions into a small fully convolution network (FCN) with dynamic filters, whose weights and bias are predicted as mask parameters. In addition, we define a clip-level instance segmentation loss to push the predicted clip-level instance masks to maintain temporal coherence in the clip. Finally, to validate the effectiveness and generality of our CiCo strategy, we apply it to two different yet representative FiFo instance segmentation frameworks, Yolact \cite{bolya2019yolact} and CondInst \cite{tian2020conditional}, and the resulting CiCo models, namely CiCo-Yolact and CiCo-CondInst, demonstrate new state-of-the-art VIS performance. 
\section{Related work}\label{sec:related}
\vspace{-1mm}

\textbf{Image instance segmentation (IIS)}. IIS networks can be roughly categorized into bottom-up and top-down paradigms. The former ones \cite{dai2016instance,li2017fully,chen2018masklab} first perform semantic segmentation and then identify the specific locations for each instance, while the latter ones \cite{he2017mask,huang2019mask,liu2018path,bolya2019yolact,cao2020sipmask,tian2020conditional,chen2020blendmask} usually detect objects first and then segment them.
Mask R-CNN \cite{he2017mask} and its successors \cite{huang2019mask,liu2018path} are representative two-stage IIS approaches, which first generate region-of-interests (RoIs) by a region proposal network and then extract features by RoIPooling or RoIAlign to classify and segment objects in the second stage. Recently, one-stage top-down approaches such as Yolact \cite{bolya2019yolact} and CondInst \cite{tian2020conditional} partition instance segmentation into two parallel sub-tasks: generating a set of image-size mask prototypes and predicting box-level mask coefficients, and then combining them linearly to obtain the final instance masks. This eliminates the need of the second stage of IIS while keeping the balance between accuracy and speed. 

\textbf{Video instance segmentation (VIS)}. Compared with IIS, VIS takes the temporal dimension into account. Approaches \cite{yang2019video,cao2020sipmask,QueryInst,liu2021sg} partition the video sequence into individual frames, and detect and segment objects frame by frame via IIS methods, finally associate objects via the object embedding vectors predicted by the newly added tracking branch. Besides, MaskProp \cite{bertasius2020classifying} and Propose-Reduce \cite{lin2021video} extend Mask R-CNN \cite{he2017mask} by introducing a branch to align the predicted frame-level instance masks from a key frame to the reference frames. The difference between them lies in that the former propagates instance masks from one frame to all the other adjacent frames, while the latter propagates instance masks from several selected key frames to all the other frames. The most commonly used techniques for mask alignment include deformable convolution \cite{bertasius2020classifying,qi2021occluded,dai2017deformable}, non-local block \cite{lin2021video}, attention \cite{fu2020compfeat,pcan}, correlation \cite{Li_2021_CVPR,qi2021occluded} and graph neural network \cite{wang2021end}, {\it etc}. A spatio-temporal embedding scheme is proposed in STEm-Seg \cite{Athar_Mahadevan20ECCV}, while the bottom-up paradigm leads to much lower performance. Inspired by the attention mechanism \cite{vaswani2017attention}, VisTR \cite{wang2020end} adds a Transformer \cite{vaswani2017attention} to generate global spatio-temporal attention, and predicts instance masks on the entire video. To alleviate heavy computational and storage overhead in VisTR, IFC \cite{hwang2021video} adopts separated spatial and temporal attentions on video clips. In this work, we propose a general strategy to extend frame-in frame-out one-stage VIS methods to clip-in clip-out ones.

\vspace{-1mm}
\section{Approach} 
\vspace{-1mm}
In this section, we first briefly introduce the general frame-in frame-out (FiFo) one-stage instance segmentation framework in Sec. \ref{sec:fifo}, then analyze the short-term temporal coherence of videos in Sec. \ref{sec:temporalcohrence} and propose the clip-in clip-out (CiCo) VIS approach in Sec. \ref{sec:cico}, finally present the overall architecture and loss function of CiCo in Sec. \ref{sec:entire_arch}. 
We denote the entire video as $\mathcal{V}=[I^1, \cdots, I^L] \in R^{L\times H\times W\times 3}$, where $H, W, L$ represent its height, width and the total number of frames, and $I^t$ is a single frame within it. Suppose that each video clip includes $T=2\delta+1$ consecutive frames, represented as $\mathcal{V}^l=[I^{t-\delta}, \cdots, I^{t+\delta}] \in R^{T\times H\times W \times 3}$,  where $t$ is the central frame index of the clip. 
 
\vspace{-1mm}
\subsection{Frame-in Frame-out One-stage VIS}\label{sec:fifo}
\begin{figure*}[!t]
\centering
\includegraphics[width=.99\textwidth]{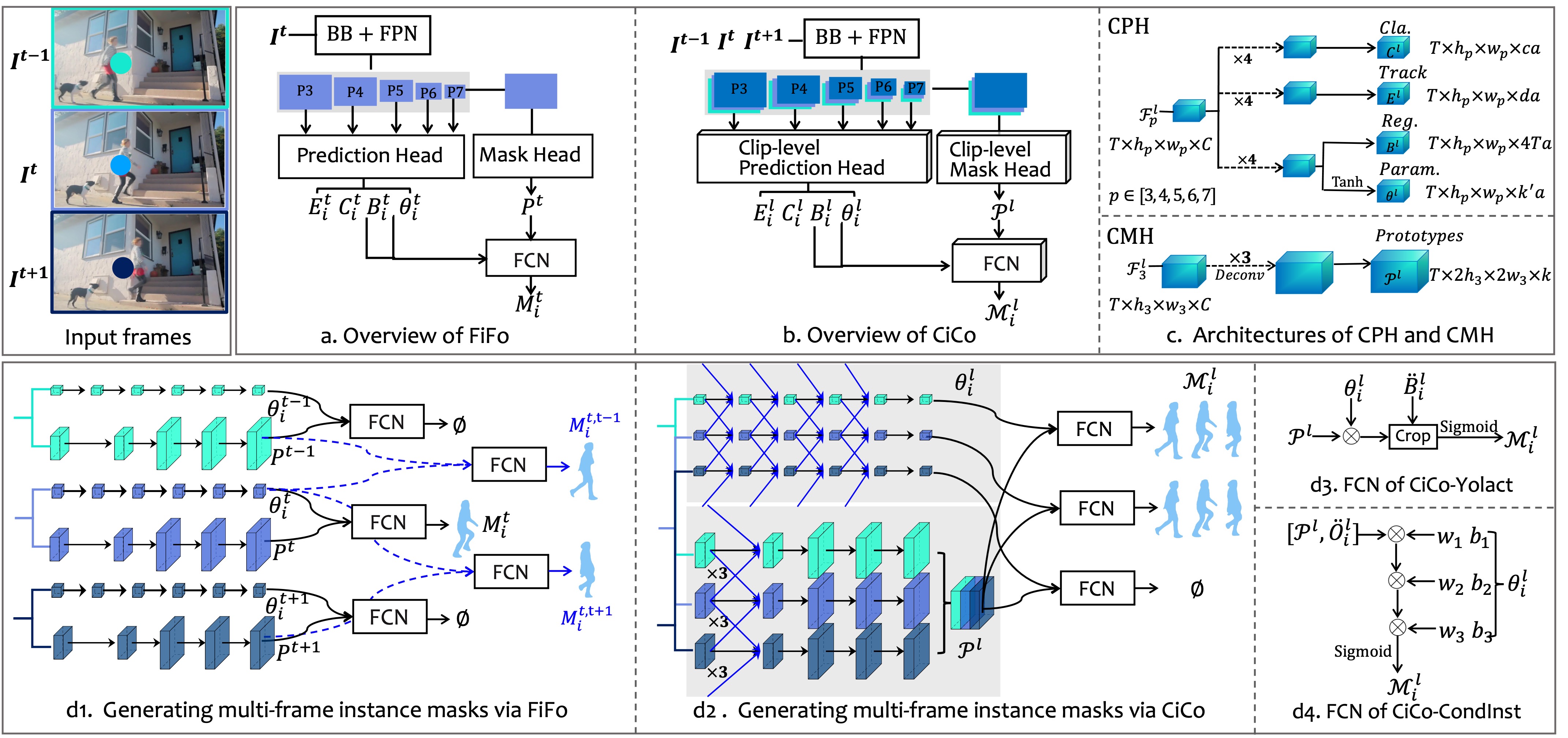}
\vspace{-3mm}
\caption{
Overviews of (a) frame-in frame-out (FiFo) and (b) clip-in clip-out (CiCo) one-stage VIS frameworks. The FiFo pipeline partitions a video sequence into individual frames to generate desired instance masks frame-by-frame, as shown in sub-figure (d1), while CiCo pipeline introduces 3D {\it Conv} into prediction heads (PH) and mask heads (MH) to extract spatio-temporal features, termed as clip-level prediction heads (CPH) and clip-level mask heads (CMH), whose architectures are shown in sub-figure (c). Sub-figure (d2) illustrates the process of generating clip-level instance masks via CiCo, and sub-figures (d3) and (d4) show the difference of FCN architectures between two representative one-stage frameworks: Yolact and CondInst. 
}\label{fig:fifo_cico}
\vspace{-3mm}
\end{figure*}

As shown in Fig. \ref{fig:fifo_cico}(a), the FiFo one-stage VIS framework inputs a single frame $I^t$ to the backbone and FPN to obtain multi-scale feature maps \{P3, P4, P5, P6, P7\}. First, the feature maps of each FPN layer will be input into the prediction heads to obtain box-level predictions, including class confidences, regression coordinates, and mask parameters. Then the feature maps of FPN layers \{P3, P4, P5\} are passed through a 2D {\it{conv}} layer and upsampled to the same resolution as P3 layer, the sum of which will be fed to the mask head to generate image-level feature maps (called prototypes in \cite{bolya2019yolact}). Then the frame-level mask can be generated by feeding the box-level predictions and image-level prototypes into a compact fully convolutional network (FCN) with dynamic convolutions, whose weights and bias are predicted as mask parameters.

The current one-stage instance segmentation approaches differ mainly in FCN architecture and instance location embedding. Specifically, Yolact \cite{bolya2019yolact} adopts the linear combination of box-level mask parameters $\theta_i^t\in R^{k'}$ and image-level prototypes $P^t\in R^{2h_3\times 2w_3\times k}$, which is equivalent to a $1\!\times\!1$ dynamic {\it{conv}} layer without bias. Here the numbers of box-level mask parameters and the channel of image-level prototypes are the same. The process can be instantiated as
\vspace{-1mm}
\begin{equation}\label{eq:yolact}
\vspace{-1mm}
{M_i^t}_{\text{Yolact}} = \text{Crop}(\sigma(P^t \theta_i^t),\ B_i^t), 
\end{equation}
where $B_i^t\in R^4$ is bounding boxes, and $\sigma$ is the Sigmoid activation function. '\text{Crop}' means that the instance mask is cropped by its bounding boxes. 
While CondInst \cite{tian2020conditional} employs three $1\!\times \! 1$ dynamic {\it{conv}} layers to form the compact '$\text{FCN}$', its input is the concatenation of the image-level prototypes $P^t$ with $k\!=\!8$ channels and a map of relative coordinates $O_i^t\in R^{2h_3\times 2w_3 \times 2}$ from all the locations on $P^t$ to the central point $(x_i^t, y_i^t)$ of its bounding box.
Here the box-level mask parameters $\theta_i^t$ include the weights and bias of all the three filters in 'FCN', and $k'\!=\!169$. The process of instance segmentation can be formulated as
\vspace{-1mm}
\begin{equation}\label{eq:condinst}
\vspace{-1mm}
{M_i^t}_{\text{CondInst}} = \sigma(\text{FCN}([P^t, O_i^t],\ \theta_i^t)).
\end{equation}

As shown in Fig. \ref{fig:fifo_cico} (d1), for a given 3-frame clip, the FiFo pipeline generates instance masks frame by frame, which can only classify, detect and segment objects based on the spatial information of a single frame. Unlike static image segmentation, more complex problems will be encountered in video instance segmentation, such as uncommon camera-to-object view, motion blur, occlusion, out of focus, {\it etc}. The FiFo methods are difficult to handle these challenging cases, resulting in many missed objects, inaccurate masks, and incorrect classifications.

\vspace{-1mm}
\subsection{Temporal Coherence} \label{sec:temporalcohrence}
\begin{figure}[t]
\includegraphics[width=1\linewidth]{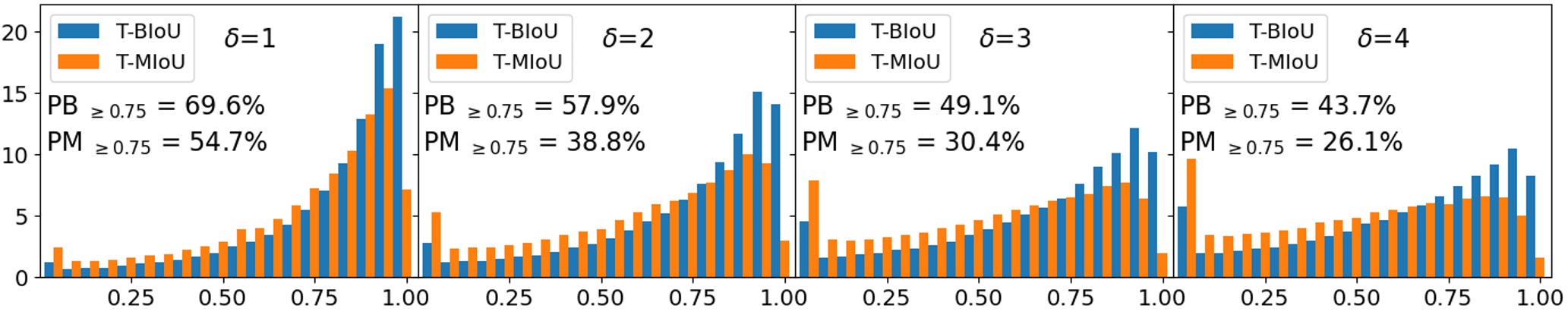}
\vspace{-6mm}
\caption{
Histograms of T-BIoU and T-MIoU on objects between two adjacent frames with $\delta$-frame interval on the YouTube-VIS 2019 train set. 'PB$_{\geq 0.75}$' and 'PM$_{\geq 0.75}$' denote the proportions of objects with T-BIoU and T-MIoU greater than 0.75.}
\label{fig:TIoU}
\vspace{-2mm}
\end{figure}

The fact that the contents of adjacent frames in a short time are highly coherent inspires us to perform VIS clip by clip. To quantify temporal coherence of objects in a video, we define the temporal bounding-boxes IoU (T-BIoU) and temporal masks IoU (T-MIoU) of an object in two adjacent frames as follows: 
\vspace{-2mm}
\begin{equation}
{\textit{T-BIoU}} = \frac{B_i^{t} \cap B_i^{t+\delta}}{B_i^{t} \cup B_i^{t+\delta}}, \quad 
{\textit{T-MIoU}} = \frac{M_i^{t} \cap M_i^{t+\delta}}{M_i^{t} \cup M_i^{t+\delta}}.
\vspace{-1mm}
\end{equation}
where $B_i^t, B_i^{t+\delta}$ and $M_i^t, M_i^{t+\delta}$ are the bounding boxes and masks of $i$-th instance in frames $t$ and $t+\delta$, respectively, and $\delta$ is the number of interval frames. For an instance in frame $t$ of a video, if the object appears in the $t-\delta$ and $t+\delta$ frames at the same time, its T-BIoU and T-MIoU are calculated on both frames and then averaged. Otherwise, only the frame where it appears is used in computation.

After traversing all videos in the YouTube-VIS 2019 train set, we plot the histograms of T-BIoU and T-MIoU in Fig. \ref{fig:TIoU}, where $\delta$ is from 1 to 4 and the bin step is 0.05. The proportions of objects with T-BIoU and T-MIoU greater than 0.75, denoted by PB$_{\geq 0.75}$ and PM$_{\geq 0.75}$, are also shown in the sub-figures. Surprisingly, the proportions PB$_{\geq 0.75}$ and PM$_{\geq 0.75}$ reach almost 70\% and 55\% between two frames with $\delta\!=\!1$, and 58\% and 39\% between two frames with $\delta = 2$. In other words, if the video can 'see' an object at a specific location in a frame, there is a high probability that the video can 'see' the same object at the same location in adjacent frames. Therefore, it is feasible to employ naive 3D {\it{conv}} layers to obtain spatio-temporal features in a short video clip for object detection and segmentation. Fig. \ref{fig:TIoU} also shows that with the increase of $\delta$, the proportions PB$_{\geq 0.75}$ and PM$_{\geq 0.75}$ drop a lot, especially for T-MIoU. Indeed, the temporal coherence decreases with the increase of time interval in a video.  

\vspace{-1mm}
\subsection{Clip-in Clip-out One-stage VIS} \label{sec:cico}
The strong temporal coherence in a clip implies that even the naive 3D {\it{conv}} layer can be used to effectively extract the spatio-temporal semantic information of objects. As shown in Fig. \ref{fig:fifo_cico}(b), to endow FiFo based one-stage framework the capability of generating instance masks clip by clip without much additional storage overhead, we keep the backbone network and FPN unchanged to extract multi-scale feature maps and only replace 2D {\it{conv}} layers of the prediction and mask heads with 3D {\it{conv}} layers to extract spatio-temporal information, which are namely as clip-level prediction heads (CPH) and clip-level mask head (CMH). 
% On each feature scale, feature maps of all frames in the clip are stacked to form a feature cube for subsequent use, denoted as $\mathcal{F}_p^l \in R^{T\times h_p\times w_p\times C}$, where $p \in \{3,4,5,6,7\}$. 

\textbf{Clip-level prediction heads (CPH).}
As shown in Fig. \ref{fig:fifo_cico}(c), CPH consists of three branches to predict $c$ class confidences, $d$ embedding vectors of tracking, $4T$ bounding box regression coordinates and $k'$ mask parameters, and each branch is composed of a prediction tower with four {\it{conv}} layers and a final prediction layer. Since object bounding box regression and mask parameter prediction are location-sensitive, they employ a shared prediction tower to maintain feature consistency. Benefiting from the short-term temporal coherence, the box-level predictions of an object throughout a short video clip maintain the {\it{temporal consistency}}. Thus, we can adopt shared box-level predictions for an object throughout the short video clip. 
For these branches in CPH with 3D {\it{conv}} layers, we stack the FPN features of all frames in a clip in temporal dimension to form a spatio-temporal feature cube as input, otherwise directly stack them in the batch dimension. Sec. \ref{sec:abl} provides more discussions and performance comparisons on how to introduce 3D {\it Conv} into different branches of CPH.

\textbf{Clip-level mask head (CMH).}
The architecture of CMH is shown at the bottom of Fig. \ref{fig:fifo_cico}(c), which only inflates the original mask head by introducing 3D {\it{conv}} layers. CMH includes three 3D {\it{conv} } layers with $3\times 3 \times  3$ filters to embed spatio-temporal features, a deconvolutional layer for upsampling features to 1/4 resolution of the input frames, a $1\times 3 \times  3$ and a $1\times 1\times 1$ 3D {\it{conv} } layers to reduce channels from $C$ to $k$. The input of CMH is the sum of multi-scale feature maps from \{P3, P4, P5\} FPN layers after convolution and upsampling operations, whose shape is the same as the resolution of P3 FPN layer, and its output is a set of clip-level mask features $\mathcal{P}^l \in R^{T\!\times\! 2h_3\! \times\! 2w_3\! \times\! k}$, called clip-level prototypes. Due to the temporal coherence, the clip-level prototypes are highly {\it{structural homogeneous}} along with the temporal dimension.

\textbf{Clip-level mask generation.}
Since the bounding boxes of an object overlap with each other severely in frames of a clip and they are mainly used to locate instances, it is a simple and feasible choice to replace bounding boxes in each single frame with their corresponding circumscribed boxes throughout the clip. 
We denote the bounding boxes of the $i$-th object in frame $t$ as $B_i^t=[{x}_{i1}^t, {y}_{i1}^t, {x}_{i2}^t, {y}_{i2}^t]$, whose first and last two elements are the coordinates of the upper-left pixel and bottom-right pixel. Its circumscribed boxes $\ddot{B}_i\in R^4$ can be computed by:
\vspace{-1mm}
\begin{equation}\label{eq:cir_box1}
\ddot{B}_i^l = [\min_{t}\{{x}^t_{i1}\}, \min_t\{{y}^t_{i1}\}, \max_{t}\{{x}^t_{i2}\}, \max_t\{{y}^t_{i2}\}],\ \forall t \in [t-\delta, t+\delta].
\vspace{-1mm}
\end{equation}
Then, the clip-level instance masks $\mathcal{M}^{l}_i$ of the $i$-th instance in the $l$-th clip can be obtained by inputting box-level predictions and the clip-level prototypes $\mathcal{P}^l$ into a compacted FCN with dynamic convolutions, whose weights and bias constitute mask parameters $\theta_i^l$. Note that all dynamic convolutions in FCN are $1\times 1 \times 1$ filters.
Specifically, by extending the FCN architecture of FiFo approach Yolact \cite{bolya2019yolact} (refer to Eq. (\ref{eq:yolact})), the clip-level instance masks of the $i$-th instance with $T\times 2h_3\times 2w_3$ shape, called CiCo-Yolact, can be formulated as follows:
\begin{equation}\label{eq:yolact_cico}
{\mathcal{M}_i^l}\ _{\text{Yolact}} = \text{Crop}(\sigma(\mathcal{P}^l \theta_i^l), \ddot{B}_i^l)).
\vspace{-1mm}
\end{equation}
While, by extending FiFo based approach CondInst \cite{tian2020conditional} (refer to Eq. (\ref{eq:condinst})), the clip-level instance masks, called CiCo-CondInst, can be formulated as follows:
\begin{equation}\label{eq:condinst_cico}
\vspace{-1mm}
\mathcal{M}_i^l\ _{\text{CondInst}} = \sigma(\text{FCN}([\mathcal{P}^l, \ddot{O}_i^l],\ \theta_i^l)),
\end{equation}
where $\ddot{O}_i^l\in R^{T\times 2h_3\times 2w_3 \times 2}$ means a map of relative coordinates from all the locations on $\mathcal{P}^l$ to the central point $(\ddot{x}_i^l, \ddot{y}_i^l)$ of its clip-level circumscribed bounding box.
Figs. \ref{fig:fifo_cico} (d3) and (d4) show the schematic diagrams of clip-level instance mask generation in CiCo-Yolact and CiCo-CondInst, respectively.

\textbf{Sample matcher.}
Several anchors with different aspect ratios and scales are predefined at each position of an FPN feature on anchor-based detectors. 
During training, if the IoU between an anchor and a ground-truth bounding box of an object is greater than a preset threshold $\epsilon_p$ (e.g., 0.5), the corresponding anchor is called a positive sample. If the IoU between an anchor and all ground-truth bounding boxes of all objects is lower than a certain threshold $\epsilon_n$ (e.g.,  0.4), the anchor will be assigned to background, called a negative sample. As analysed in Sec. \ref{sec:temporalcohrence}, the contents of adjacent frames are highly coherent, and the object displacements in two adjacent frames are usually small. Utilizing the circumscribed boxes of objects in adjacent frames to match the positive and negative samples can select appropriately larger anchors to reach richer spatio-temporal features. Meanwhile, since objects may have larger motions as the frame interval increases, we only adopt object bounding boxes of the current frame, previous frame, and next frame to calculate their circumscribed boxes. 

\subsection{Overall Architecture} \label{sec:entire_arch}
\textbf{Training.} We consecutively sample $\delta$ frames before and after the query frame, forming a clip of $2\delta+1$ frames. We first input it into CiCo to obtain the box-level predictions from CPH and the clip-level prototypes from CMH as described in Sec. \ref{sec:cico}, and match positive and negative samples by the circumscribed boxes. 
We adopt the smooth $L_1$ loss ($SL_1$) and the cross entropy ($CE$) loss for bounding box regression ${B}^l_i\! \in\! R^{4T}$ and classification ${C}^l_i\! \in\! R^{c}$ respectively, as follows:
\vspace{-1mm}
\begin{equation}
\vspace{-1mm}
\mathcal{L}_{cls}\!=\! \frac{1}{N^l} \sum\nolimits_{i=1}^{N^l} CE(\bar{C}^l_i, C^l_i),\quad
\mathcal{L}_{reg}\!=\! \frac{1}{N^l} \sum\nolimits_{i=1}^{N^l} SL_1(\bar{B}^l_i, B^l_i),
\end{equation}
where $N^l=N^l_{pos}\!+\! N^L_{neg}$ is the total number of positive and negative samples, $\bar{*}$ represent the corresponding ground-truths items.
For the positive samples, we obtain the desired clip-level instance segmentation by Eq. (\ref{eq:yolact_cico}) or (\ref{eq:condinst_cico}). The clip-level instance segmentation loss is based on the binary cross entropy ($\text{BCE}$) loss:
\vspace{-1mm}
\begin{equation}
\mathcal{L}_{mask}  = \frac{1}{N^l_{pos}} \min_{\mathcal{P}^l}  \sum\nolimits_{i=1}^{N^l_{pos}} \min_{\theta^l_i} \text{BCE}( \bar{\mathcal{M}}_i^l,\ \text{FCN}(\mathcal{P}^l,\ \theta^l_i, \ddot{B}^l_i)). 
\end{equation}
Based on the cosine similarity of embedding vectors of any two positive samples $\ d_{ij} = 0.5(\frac{E_i^l\cdot E_j^l} {\parallel E_i^l\parallel \parallel E_j^l \parallel} + 1)\in [0, 1]$. The tracking loss can be written as: 
\vspace{-1mm}
\begin{equation}
\mathcal{L}_{track} = -\frac{1}{(N^l_{pos})^2} \sum\nolimits_{i,j=1}^{N_{pos}^l}  \log d_{ij}|_{\text{ID}_i = \text{ID}_j} +\log(1-d_{ij})|_{\text{ID}_i \neq \text{ID}_j}. 
\vspace{-1mm}
\end{equation}

Finally, the total loss function of our proposed CiCo for one-stage VIS is:
\vspace{-1mm}
\begin{equation}
\vspace{-1mm}
    \mathcal{L}_{total}  = \lambda_1\mathcal{L}_{cls} + \lambda_2 \mathcal{L}_{reg} + \lambda_3\mathcal{L}_{mask} + \lambda_4\mathcal{L}_{track}. 
\end{equation}
where $\lambda_1, \lambda_2, \lambda_3, \lambda_4$ are balance parameters.

\textbf{Inference.} During inference, we partition the input video into several clips with an interval of $2\delta+1$ frames. All frames of each clip are input to CiCo to obtain the box-level predictions and clip-level prototypes. In detection, we filter out those objects with class confidence below 0.1, and generate the clip-level instance segmentation by Eq. (\ref{eq:yolact_cico}) or (\ref{eq:condinst_cico}) for the remaining objects. After sorting them by class confidence, we employ NMS to remove duplicate objects, and link objects across clips according to the following clip-level online tracking strategy. 

\textbf{Clip-level tracking.} A naive clip-level online tracking strategy is used during inference. For the first clip, we assign the IDs of all detected instances as $\mathcal{Y}_{\text{ID}}=\{1,...,N^1\}$, where $N^1$ is the number of detected instances in the first clip. In the following clips, we associate predicted objects in the current clip based on the matching scores of objects between the previous clip and the current clip. The matching score combines the cosine similarity of embedding vectors, the mask IoU ($\text{MIoU}$) and bounding box IoU ($\text{BIoU}$) of objects, formulated as
\vspace{-1mm}
\begin{equation}
\begin{aligned}
s_{i,j}^{l\!-\!1,l} = &\ \alpha_1 \text{d}(E_i^{l-1}, E_j^l) + \alpha_2 \text{MIoU}(\mathcal{M}_i^{l-1},\ \mathcal{M}_j^{l}) + \alpha_3 \text{BIoU}(B_i^{l-1},\ B_j^{l}), 
\vspace{-1mm}
\end{aligned}
\end{equation}
where $i, j$ denote instance IDs from the ($l\!-\!1$)-th clip and the $l$-th clip, $\alpha_1, \alpha_2, \alpha_3$ are the hyperparameters. For the $j$-th instance in the current clip, if its highest matching score computed with all objects in the previous clip is greater than a certain threshold, it is assigned with the ID of that instance in the previous clip. Otherwise, it is considered as a new one with ID $|\mathcal{Y}_{\text{ID}}|\!+\!1$.

\section{Experimental Results}

% \subsection{Implementation details}
\textbf{Datasets and evaluation metrics.} YouTube-VIS \cite{yang2019video} 2019 and 2021 respectively contain 2,283 and 2,985 training, 302 and 421 validation, and 343 and 453 test videos. All the videos are annotated for every 5 frames. The number of frames per video is between 19 and 36. Both the two datasets have 40 common object categories, but there are some differences in the categories. OVIS \cite{qi2021occluded} includes 607 training, 140 validation and 154 test videos, scoping 25 object categories. Different from YouTube-VIS 2019 or 2021, videos in OVIS are longer and have more objects per frame. The commonly used metrics, including average precision (AP) at different IoU thresholds, average recall (AR) and the mean value of AP (mAP), are adopted for VIS model evaluation.

\textbf{Model setup.} We adopt ResNet 50/101 \cite{he2016deep} with a feature pyramid network (FPN) \cite{lin2017feature} and Swin Transformer tiny (SwinT-tiny) \cite{Liu_2021_ICCV} as backbones, which are initialized by pre-trained models on MS-COCO datasets \cite{lin2014microsoft}. Applying CiCo to two representative one-stage FiFo instance segmentation baselines, Yolact \cite{bolya2019yolact} and CondInst \cite{tian2020conditional}, we instantiate two models of CiCo, termed as CiCo-Yolact and CiCo-CondInst, both of which employ the anchor-based detector for simplicity. Following the settings in \cite{yang2019video}, all frames are resized to $360\times 640$ during training and inference. All models are trained on four V100 GPUs with 32G RAM and tested on a single Nvidia 2080Ti. The training epoch is 12 and the batch sizes are 24 clips for all datasets. The learning rate starts at 0.001 and decays by 0.1 at epochs 8 and 10. It is worth mentioning that when 3-frame clips are used as the input, CiCo-Yolact and CiCo-CondInst cost only 17 GB memory and 18 hours to train 12 epochs.

\subsection{Ablation study}\label{sec:abl}
We design a series of ablation experiments to verify the effectiveness of our proposed CiCo one-stage VIS approach by using ResNet50 backbone. 
\begin{figure*}[!t]
\begin{minipage}[h]{1\textwidth}
\begin{minipage}[h]{0.42\textwidth}
\centering
\setlength{\tabcolsep}{0.4mm}{
      \linespread{2}
      \begin{tabular}{p{0.08\textwidth}<{\centering}p{0.11\textwidth}<{\centering}p{0.13\textwidth}<{\centering}p{0.135\textwidth}<{\centering}p{0.135\textwidth}<{\centering}p{0.13\textwidth}<{\centering}p{0.14\textwidth}<{\centering}}
         \Xhline{0.8pt}
          $T$ & Loss &  AP & AP$_{50}$  & AP$_{75}$ & AR$_{10}$ & FPS\\
          \Xhline{0.5pt}
          1& $\times$  & 30.7 & 47.8 & 31.3 & 38.0 & 23.5\\
          2& \checkmark& 33.0 & 51.6 & 35.9 & 41.0 & 23.3\\
          3& \checkmark& 32.7 & 50.2 & 35.4 & 40.3 & 23.2\\
          4& \checkmark& 29.5 & 46.4 & 30.7 & 36.9 & 23.2\\
         \Xhline{0.8pt}
      \end{tabular}
}\\
\vspace{+.5mm}
(a) Clip-level segmentation loss \\
\vspace{+1mm}
\end{minipage}
\hfill
\begin{minipage}[h]{0.55\textwidth}
\centering
\setlength{\tabcolsep}{0.4mm}{
      \linespread{2}
      \begin{tabular}{p{0.05\textwidth}<{\centering}|p{0.08\textwidth}<{\centering}p{0.08\textwidth}<{\centering}p{0.08\textwidth}<{\centering}|p{0.08\textwidth}<{\centering}p{0.1\textwidth}<{\centering}p{0.1\textwidth}<{\centering}p{0.085\textwidth}<{\centering}p{0.1\textwidth}<{\centering}|p{0.09\textwidth}<{\centering}}
         \Xhline{0.8pt}
          $T$ & Box & Tra. &Cla. &  AP & AP$_{50}$  & AP$_{75}$ & AR$_1$ &AP$_{10}$ & FPS\\
          \Xhline{0.5pt}
          3&2D     & 2D &2D & 32.7 & 50.2 & 35.4 & 34.6 & 40.3 & 23.2\\
          3&3D     & 2D &2D & 35.9 & 54.5 & 39.4 & 36.0 & 43.0 & 22.9\\ %14.6 & 37.0 & 46.0 
         %   3&3D     & 3D &2D & 35.1 & 53.8 & 37.2 & 34.9 & 41.7 & 19.8\\ % 15.6 & 35.7 & 45.4
          3&3D     & 3D &2D & 36.8 & 57.3 & 40.1 & 36.4 & 43.4 & 21.8\\ %16.8 &36.9 & 46.9
          3&3D     & 3D &3D & 35.5 & 56.8 & 36.8 & 35.6 & 41.9 & 20.8\\ % 15.5 & 35.3 & 47.9 
         \Xhline{0.8pt}
      \end{tabular}
}\\
\vspace{+.5mm}
(b) 3D {\it Conv} in different branches of CPH\\
\vspace{+1mm}
\end{minipage}
\vfill
\begin{minipage}[h]{0.42\textwidth}
\centering
\setlength{\tabcolsep}{0.4mm}{
      \linespread{2}
      \begin{tabular}{p{0.075\textwidth}<{\centering}p{0.13\textwidth}<{\centering}p{0.12\textwidth}<{\centering}p{0.135\textwidth}<{\centering}p{0.135\textwidth}<{\centering}p{0.12\textwidth}<{\centering}p{0.14\textwidth}<{\centering}
      }
         \Xhline{0.8pt}
          $T$ &  Circ. & AP & AP$_{50}$& AP$_{75}$  & AR$_1$ &AR$_{10}$\ \\ 
          \Xhline{0.5pt}
          3& $\times$   & 36.8 & 57.3 & 40.1 & 36.4 & 43.4 \\
          3& \checkmark & 37.1 & 57.4 & 40.4 & 35.6 & 43.3\\
          7& $\times$   & 33.8 & 50.6 & 38.1 & 34.4 & 39.4\\
          7& \checkmark & 35.6 & 56.5 & 38.2 & 36.9 & 43.4\\
         \Xhline{0.8pt}
      \end{tabular}
}\\
\vspace{+.3mm}
(c) Sample matcher\\
\end{minipage}
\hfill
\begin{minipage}[h]{0.55\textwidth}
\centering
\setlength{\tabcolsep}{0.4mm}{
      \linespread{2}
      \begin{tabular}{p{0.055\textwidth}<{\centering}p{0.12\textwidth}<{\centering}p{0.09\textwidth}<{\centering}p{0.1\textwidth}<{\centering}p{0.1\textwidth}<{\centering}p{0.1\textwidth}<{\centering}p{0.085\textwidth}<{\centering}p{0.095\textwidth}<{\centering}p{0.095\textwidth}<{\centering}p{0.11\textwidth}<{\centering}}
         \Xhline{0.8pt}
          $T$ &  CMH &  AP & AP$_{50}$\  & AP$_{75}$ & AP$_s$ &AP$_m$& AP$_l$ &FPS\\
          \Xhline{0.5pt}
          3&2D  & 36.8 & 57.3 & 40.1 & 16.8 &36.9 & 46.9 &21.8\\
          3&3D  & 37.1 & 55.2 & 39.7 & 14.3 &41.0 & 48.4 &20.7\\
          4&2D  &36.1 & 54.1 & 40.2 & 10.3 & 36.2 & 49.5 &22.1\\
          4&3D  &36.6 & 55.4 & 39.6 & 14.2 & 39.7 & 47.5  &21.0\\
        %   5&2D\\
        %   5&3D\\
         \Xhline{0.8pt}
      \end{tabular}
}\\
\centering
\vspace{+0.5mm}
(d) 3D {\it Conv} in CMH\\
\end{minipage}
\vspace{-2.5mm}
\tabcaption{ Ablation study of CiCo-Yolact with ResNet50 backbone on YouTube-VIS 2019 valid set. Sub-tables (a) and (b) study the clip-level instance segmentation loss under different clip lengths by progressively introducing 3D {\it Conv} into branches of CPH. Sub-table (c) tests the sample matcher with object bounding boxes or circumscribed boxes, and sub-table (d) compares CMH with 2D and 3D {\it Conv}.
}\label{tab:abl}
\end{minipage}
\vspace{+2mm}
% ------------------------------------------------------------------
\vfill
\begin{minipage}[h]{1\textwidth}
\centering
\includegraphics[width=0.99\linewidth]{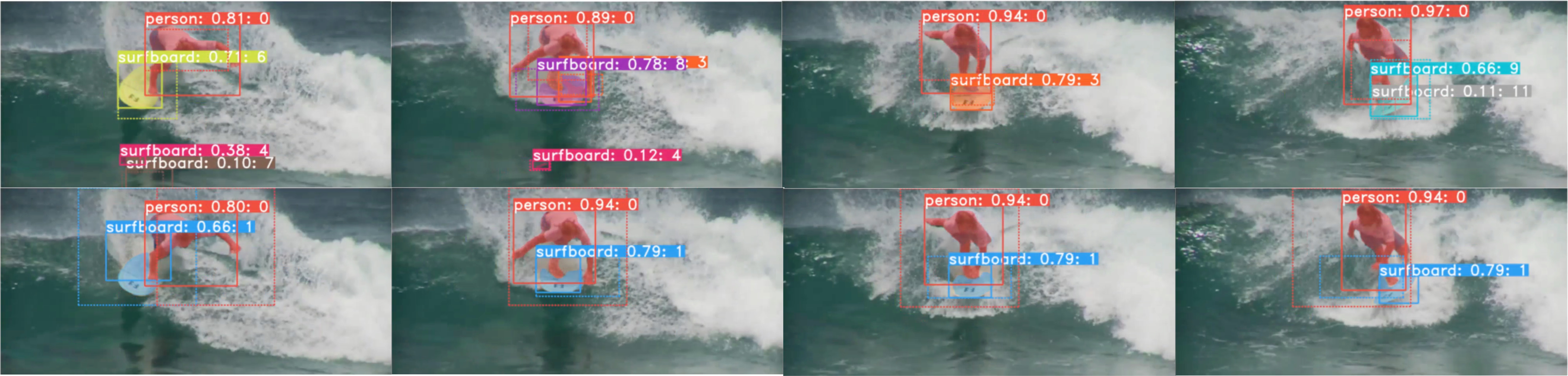}
\vspace{-3.5mm}
\caption{Visualization results of Yolact (top row) and CiCo-Yolact (bottom row) with a 3-frame clip. The dashed and solid boxes are anchors and predicted bounding boxes.}\label{fig:vis_masks}
\end{minipage}
\vspace{+2mm}
% ------------------------------------------------------------------
\vfill
\begin{minipage}[h]{0.42\textwidth}
\centering
\setlength{\tabcolsep}{0.5mm}{
      \linespread{2}
      \begin{tabular}{p{0.16\textwidth}<{\centering}p{0.09\textwidth}<{\centering}p{0.14\textwidth}<{\centering}p{0.155\textwidth}<{\centering}p{0.155\textwidth}<{\centering}p{0.155\textwidth}<{\centering}}
          \Xhline{0.8pt}
          Data &$T_{o}$ &  AP    &AP$_{50}$ & AP$_{75}$  & AR$_{10}$\\
          \Xhline{0.5pt}
          \multirow{3}{*}{YT21}
          & 0 & 33.5 & 51.9 & 36.4 &39.2\\
          & 1 & 34.3 & 55.0 & 36.8 &39.5\\
          & 2 & 34.8 & 53.7 & 38.9 &39.7\\
          \Xhline{0.5pt}
          \multirow{3}{*}{OVIS}
          & 0 & 14.8 & 31.7 & 13.7 & 19.3\\
          & 1 & 17.2 & 35.5 & 15.8 & 23.0\\
          & 2 & 17.4 & 34.5 & 17.2 &22.9\\
          \Xhline{0.8pt}
      \end{tabular}
}
\vspace{-3.5mm}
\tabcaption{ Ablation study of CiCo-Yolact with different overlap frames ($T_0$). The inputs are 3-frame clips.}\label{tab:abl_over}
\end{minipage}
\hfill
\begin{minipage}{0.54\textwidth}
\centering
\vspace{-1mm}
\includegraphics[width=0.93\linewidth]{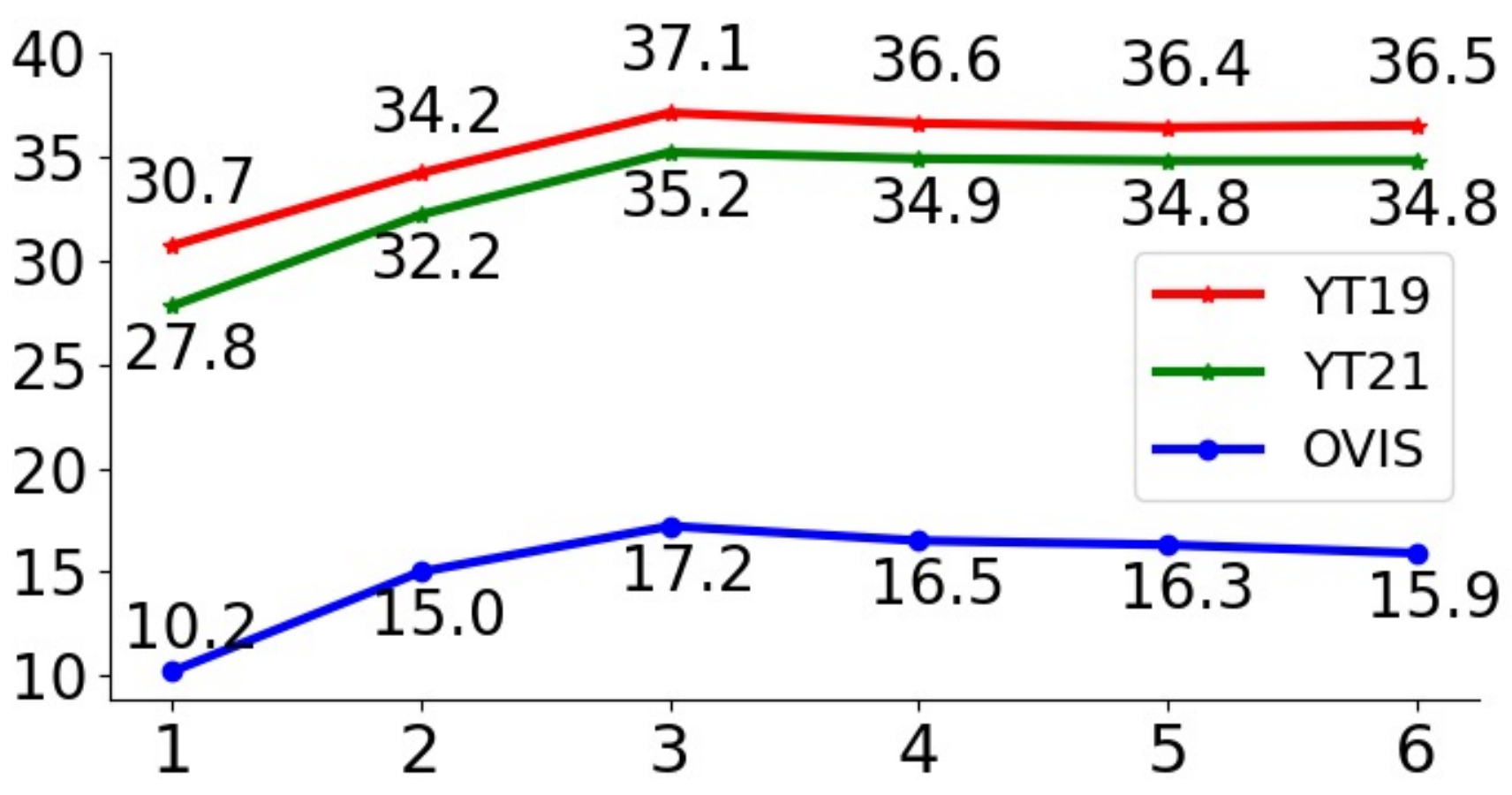}\\
\vspace{-3mm}
\caption{CiCo-Yolact with ResNet50 backbone under different clip lengths on benchmarks. }\label{fig:abl_benchmark}
\end{minipage}
\vspace{-2mm}
\end{figure*}

\textbf{Clip-level segmentation loss.}
Table \ref{tab:abl}(a) reports the results of Yolact by using clip-level instance segmentation loss. It obtains box-level predictions from PH and prototypes from MH frame by frame. For each object, the clip-level instance segmentation loss includes two parts: intra-frame and inter-frame segmentation loss. The former is the original frame-level instance segmentation loss, while the latter forces instance masks to keep consistency between adjacent frames. We directly input mask parameters of the current frame and prototypes of the adjacent frames into FCN to generate shifted instance masks (please refer to the blue dotted lines in Fig. \ref{fig:fifo_cico}(d1)). Compared with single frame segmentation, the above naive clip-level instance segmentation loss on 3-frame clips can lead to about 3\% mask AP performance improvement. However, as the number of frames increases, the temporal coherence decreases rapidly, resulting in performance degradation since the features are still extracted from the current single frame.

\textbf{CPH.} 
Table \ref{tab:abl}(b) shows a comprehensive ablation study on CPH. Compared with the FiFo baseline (first row), when employing only 3D {\it Conv} in bounding boxes prediction tower to extract spatio-temporal features to predict bounding box regression and mask parameters (second row), CiCo-Yolact can directly gain 3.2\% mask AP, improving from 32.7\% to 35.9\%. After replacing 2D {\it Conv} with 3D {\it Conv} in the tracking prediction tower, the performance can be further improved to 36.8\%. However, introducing 3D {\it Conv} in the classification prediction tower degrades performance, because the 3D {\it Conv} mixes different levels of background information depending on the motion of objects, which may confuse the classifier. Thus, we adopt 3D {\it Conv} layers for the bounding boxes and tracking prediction towers, and adopt 2D {\it Conv} layers for the classification prediction tower.

\textbf{Circumscribed boxes of sample matcher.} In Table \ref{tab:abl}(c), we test the use of circumscribed boxes in sample matcher on different clip lengths. We see that compared with using bounding boxes, using circumscribed boxes can maintain equally good performance on shorter clips (i.e.,  3-frame clips) but achieve higher performance on longer clips (i.e., 7-frame clips). This is because shorter clips have stronger temporal coherence within frames, but longer clips require the help of appropriate width of boxes to take advantage of a larger receptive field.

\textbf{CMH.} 
Table \ref{tab:abl}(d) compares 2D {\it Conv} with 3D {\it Conv} in CMH. We see that 3D {\it Conv} brings 0.3\% overall improvement. In specific, 3D {\it Conv} increases the performance significantly on medium and large objects, but drops slightly on small objects because small objects are more likely to have faster motion.

\textbf{Visual comparison of instance masks.}
Fig. \ref{fig:vis_masks} compares the instance masks results produced by FiFo based Yolact and CiCo-Yolact. We see that CiCo can keep the temporal consistency of objects in category, bounding boxes, masks and tracking ID, and it can adaptively match wider anchors according to the object position in adjacent frames (e.g., 'surfboard'). The richer spatio-temporal semantic information in CiCo also reduces some false positive samples, such as the purple 'surfboard' in the first two frames of the top row.

\textbf{Overlap frames between adjacent clips.}
Table \ref{tab:abl_over} compares the VIS results of CiCo-Yolact by using different number of overlap frames between two adjacent clips, where the clip length is 3. On YouTube-VIS 2021 and OVIS valid sets, compared with no-overlap (i.e., $T_{o}=0$), 1-frame overlap significantly improves the performance by 1.1\% and 3.4\% mask AP, while 2-frame overlap further improves the mask AP by 0.5\% and 0.2\%. There is more significant performance improvement on OVIS because videos in OVIS have more heavily occluded objects and longer duration, and thus clips with overlapping frames can provide more accurate cues for tracking across clips. To keep the balance of accuracy and speed, we set the default number of overlap frames between two consecutive clips as 1 on all VIS benchmark datasets.

\textbf{Clip lengths.}
% Will update the performance here
Fig. \ref{fig:abl_benchmark} displays the performance of CiCo-Yolact with different clip lengths on all benchmark datasets. On YouTube-VIS 2019 and 2021 valid sets, the performance improves significantly as the number of frames increases in the beginning, peaking at 3-frame clips, and fluctuates within 0.5\% mask AP from 4 to 6 frames per clip. While the performance on OVIS valid set fluctuates between 15.0\% and 17.2\% mask AP from 2 to 6 frames, reaching the peak at 3-frame clips. The performance fluctuation on OVIS valid set is higher because videos in OVIS include more crowded and occluded objects, and the features at the same location of adjacent frames may come from different objects.

\begin{table*}[!thb]
\begin{center}
\setlength{\tabcolsep}{0.5mm}{
      \linespread{2}
      \begin{tabular}
      % [t]{cclcccccccccc}
      {p{0.09\textwidth}<{\centering}p{0.09\textwidth}<{\centering}p{0.2\textwidth}p{0.12\textwidth}p{0.05\textwidth}<{\centering}p{0.05\textwidth}<{\centering}p{0.06\textwidth}<{\centering}p{0.06\textwidth}<{\centering}p{0.06\textwidth}<{\centering}p{0.06\textwidth}<{\centering}p{0.06\textwidth}<{\centering}}
         \Xhline{1pt}
          Data & Type & Methods & Feat. &\ $T$\ &AP &AP$_{50}$ &AP$_{75}$&AR$_{1}$&AR$_{10}$ &FPS \\
         \Xhline{1pt}
         \multirow{18}{*}{\shortstack{YT19}} & \multirow{10}{*}{\shortstack{Frame-\\ level}}
         &MaskTrack \cite{yang2019video}\             & R50 & 2  & 30.3 & 51.1 & 32.6  &31.0 &35.5 &18.4 \\
         &&QueryInst\cite{QueryInst}                  & R50 & 2  & 34.6 & 55.8 & 36.5  &35.4 &42.4 &32.3  \\
         &&CompFeat\cite{fu2020compfeat}              & R50 & 3  & 35.3 & 56.0 &38.6 &33.1 &40.3 &-\\
         &&MaskProp* \cite{bertasius2020classifying}  & R50 & 13 & 40.0 &  -   & 42.9  &-    &- & -   \\
         &&Pro-Red* \cite{lin2021video}               & R50 & 36 & 40.4 & 63.0 & 43.8 & 41.1 & 49.7 &-   \\ 
         &&SipMask \cite{cao2020sipmask}              & R50 & 2  & 32.5 & 53.0 & 33.3  &33.5 &38.9 &30.0 \\
         &&STMask\cite{Li_2021_CVPR}                  & R50+DCN  & 2  & 33.5 & 52.1 & 36.9  &31.1 &39.2 &28.6 \\
         &&CrossVIS \cite{yang2021crossover}          & R50 & 2  &34.8 & 54.6 & 37.9 & 34.0 &39.0 &39.8 \\ 
         &&VisSTG\cite{wang2021end}                   & R50 & 2  & 35.2 & 55.7 & 38.0  &33.6 &38.5 &22.0 \\         
         &&PCAN\cite{pcan}                            & R50 & -  & 36.1 & 54.9 & 39.4 & 36.3 & 41.6 &15.0 \\
         
         \cline{2-11}
         &\multirow{8}{*}{\shortstack{Clip- \\ level}} 
         &STEm-Seg \cite{Athar_Mahadevan20ECCV}       & R50 & 8   & 30.6 & 50.7 & 33.5  &31.6 &37.1 &$<$ 5    \\
         && CiCo-Yolact                               & R50 & 3   & 37.1 & 55.2 & 39.7 & \textbf{37.3} & 43.5  &20.7  \\
         && CiCo-CondInst                             & R50 & 3   & \textbf{37.3} & \textbf{57.1} & \textbf{40.7} & 36.5 & \textbf{44.3} &19.6\\
         \cline{3-11}
         &&VisTR \cite{wang2020end}                   & R50+Att & 30  & 30.7 & 53.2 & 32.8 & 31.3 & 36.0 &30.0 \\
         &&VisTR \cite{wang2020end}                   & R50+Att & 36  & 35.6 & 56.8 & 37.0 & 35.2 & 40.2 &30.0 \\
         &&IFC\cite{hwang2021video}                   & R50+Att & 5   & 41.0 & 62.1 & 45.4 & \textbf{43.5} & \textbf{52.7}&  46.5   \\
         &&CiCo-Yolact                                & Swin-tiny & 3     & \textbf{41.8} & 63.2 & \textbf{46.3} & 39.1 & 47.2 &20.2   \\
         &&CiCo-CondInst                              & Swin-tiny & 3     & 41.4 & \textbf{64.5} & 44.5 & 38.5 & 47.2 &19.8 \\

         \Xhline{1pt}
         \multirow{10}{*}{\shortstack{YT21}} & \multirow{4}{*}{\shortstack{Frame-\\ level}}
         &MaskTrack \cite{yang2019video}\           & R50 & 2  & 28.6 & 48.9 & 29.6 &- &- &18.4  \\
         &&SipMask \cite{cao2020sipmask}            & R50 & 2  & 29.6 &48.9 & 31.1 & 28.9 & 36.1 &30.0  \\
         &&STMask \cite{Li_2021_CVPR}               & R50+DCN  & 2  & 31.1 & 50.4 & 33.5 & 26.9 & 35.6 &28.6  \\
         &&CrossVIS \cite{yang2021crossover}        & R50 & 2  & 33.3 &53.8 &37.0 &30.1 &37.6 &- \\
         \cline{2-11}
         &\multirow{6}{*}{\shortstack{Clip- \\ level}} 
         &STEm-Seg \cite{Athar_Mahadevan20ECCV}      & R50 & 8  & 26.5 & 44.4 & 27.6 & 25.5& 33.7 & $<$ 5  \\
         &&CiCo-Yolact                               & R50 & 3  & 35.2 & 54.4 & 39.0 &\textbf{31.8} & 40.9 & 20.7   \\
         &&CiCo-CondInst                             & R50 & 3  & \textbf{35.4} & \textbf{54.7} & \textbf{37.2} & 31.0 & \textbf{41.0} & 19.6   \\
         \cline{3-11}
         &&IFC\cite{hwang2021video}                  &R50+Att     & 5  & 36.6 & 57.9 & 39.3 &- &- & 46.5    \\
         &&CiCo-Yolact                               &Swin-tiny  & 3  & 38.0 & 60.5 &39.3 & 33.7 & 43.8 &20.2 \\
         &&CiCo-CondInst                             &Swin-tiny  & 3   &\textbf{39.1} & \textbf{61.3} & \textbf{42.6} & \textbf{35.3} & \textbf{45.4} &19.8\\

         \Xhline{1pt}
         \multirow{13}{*}{\shortstack{OVIS}} & \multirow{7}{*}{\shortstack{Frame-\\ level}}
         &MaskTrack \cite{yang2019video}\             & R50 & 2  & 10.8 &25.3  & 8.5  &7.9 &14.9 &17.6  \\
         &&SipMask \cite{cao2020sipmask}              & R50 & 2  & 10.2 &24.7  & 7.8  &7.9 &15.8 &27.2  \\
         &&CMaskTrack \cite{qi2021occluded}           & R50 & 11 & 15.4 &33.9  &13.1  &9.3 &20.0 &-     \\
         &&CSipMask \cite{qi2021occluded}             & R50 & 11 & 14.3 &29.9  &12.5  &9.6 &19.3 &-     \\
         &&QueryInst\cite{QueryInst}                  & R50 & 2  & 14.7 &34.7  &11.6  &9.0 &21.2 &-     \\ 
         &&STMask\cite{Li_2021_CVPR}                  & R50+DCN & 2  & 15.4 &33.9  &12.5  &8.9 &21.4 &25.7  \\
         &&CrossVIS \cite{yang2021crossover}          & R50 & 2  & 14.9 &32.7 &12.1 &10.3 &19.8 &-   \\
         \cline{2-11}
         &\multirow{6}{*}{\shortstack{Clip- \\ level}}
         &STEm-Seg \cite{Athar_Mahadevan20ECCV}        & R50 & 8  &13.8 & 32.1 & 11.9 &9.1 &20.0 &$<$ 5  \\
         &&CiCo-Yolact                                 & R50 & 3  & 17.2 & 35.5 & 15.8 &9.0 &23.0 & 19.7\\
         &&CiCo-CondInst                               & R50 & 3  & \textbf{18.0} &\textbf{37.9} & \textbf{17.2} & \textbf{10.0} & \textbf{23.8} & 19.3\\
         \cline{3-11}
         &&VisTR\dag \cite{wang2020end}                & R50+Att & 18 &10.2 & 25.7 & 7.7  &7.0 &17.4 &17.1  \\
         &&CiCo-Yolact                                 & Swin-tiny  & 3  & 18.0 & \textbf{37.5} & 16.2 & 10.3 & \textbf{24.0} & 19.4\\
         &&CiCo-CondInst                               & Swin-tiny  & 3  & \textbf{17.2}& 36.2 & \textbf{16.3} & \textbf{10.5} &23.9 & 19.2\\
        \Xhline{1pt}
      \end{tabular}
}
\end{center}
\vspace{-2mm}
\caption{Quantitative performance comparison of VIS methods on benchmarks. 
% The ResNet50 backbone is from \cite{he2016deep}. 
'Att' denotes the attention features in DETR transformer \cite{vaswani2017attention,carion2020end}. 'Swin-tiny' denotes the Swin transformer tiny backbone \cite{liu2021Swin}. 'R50+DCN' replaces some conv layers of ResNet50 with deformable convolutional layers \cite{dai2017deformable}. Results of all frame-level methods on OVIS are copied from \cite{qi2021occluded}, and VisTR\dag is trained by us on four V100 GPUs. '*' means that the authors used their own training settings mainly for competitions. The FPS results reported here include the data loading process.  
% Results with ResNet101 backbone are provided in the \textbf{supplementary materials}.
% '-' means that the results are not available.
}\label{tab:sota_VIS}
\vspace{-12mm}
\end{table*}

\subsection{Main Results with ResNet50 or Swin-tiny Backbone}
Table \ref{tab:sota_VIS} compares the proposed CiCo with state-of-the-art VIS methods on benchmark datasets. Since some recent approaches \cite{wang2020end,hwang2021video} perform clip-level VIS by introducing the DETR transformer \cite{vaswani2017attention,carion2020end} to encode deeper features, for fair comparison, apart from using ResNet50 as backbone, we also evaluate CiCo by employing the Swin-tiny \cite{liu2021Swin} as backbone to introduce self-attention blocks. 

\textbf{YouTube-VIS 2019 valid set.} MaskTrack R-CNN \cite{yang2019video} introduces a track embedding branch to the representative two-stage IIS method Mask R-CNN \cite{he2017mask}, and it only obtains 30.3\% mask AP. QueryInst \cite{QueryInst} follows the Sparse R-CNN \cite{sun2021sparse} pipeline and it obtains 34.6\% mask AP. CompFeat \cite{fu2020compfeat}, MaskProp \cite{bertasius2020classifying} and Propose-Reduce \cite{lin2021video} add inter-frame alignment and some specific training settings (mainly for competitions), reaching 35.3\%, 40.0\% and 40.4\% mask AP. MaskProp combines multiple networks as well, including hybrid task cascade network and high-resolution mask refinement. Other frame-level methods are mainly based on one-stage IIS frameworks Yolact \cite{bolya2019yolact} or CondInst \cite{tian2020conditional}. Yolact based methods SipMask \cite{cao2020sipmask} and STMask \cite{Li_2021_CVPR} achieve 32.5\% and 33.5\% mask AP, while CondInst based methods CrossVIS \cite{yang2021crossover}, VisSTG \cite{wang2021end} and PCAN \cite{pcan} can reach higher performance (34.8\%, 35.2\% and 36.1\%  mask AP). For clip-level methods, STEm-Seg \cite{Athar_Mahadevan20ECCV} with the bottom-up paradigm achieves only 30.6\% mask AP, while our CiCo-Yolact and CiCo-CondInst with 3-frame clips as input can achieve 37.1\% and 37.3\% mask AP. It is worth mentioning that CiCo-Yolact and CiCo-CondInst improve by 2.5\% mask AP over their corresponding frame-level baselines, such as SipMask and CrossVIS, without any inter-frame alignment. 

For clip-level methods with transformer features, VisTR \cite{wang2020end} employs 12 attention blocks but it obtains only 31.7\% and 35.6\% mask AP with 30-frame and 36-frame clips, respectively. It requires large memory resources to accommodate long clips and consumes long training time (about 8 days for 36-frame clip using eight V100 GPUs). To alleviate this issue, IFC \cite{hwang2021video} separates spatial and temporal attention blocks on shorter video clips, obtaining state-of-the-art performance (41.0\% mask AP). 
We replace the ResNet50 backbone with the Swin-tiny backbone to introduce 12 self-attention blocks in CiCo. The CiCo-Yolact and CiCo-CondInst with Swin-tiny backbone can achieve 41.8\% and 41.4\% mask AP, recording new state-of-the-art performance. 

\textbf{YouTube-VIS 2021 valid set.} YouTube-VIS 2021 is an updated version of YouTube-VIS 2019. The two-stage frame-level method MaskTrack R-CNN \cite{yang2019video} obtains only 28.6\% mask AP, while the one-stage frame-level methods SipMask \cite{cao2020sipmask}, STMask \cite{Li_2021_CVPR} and CrossVIS \cite{yang2021crossover} achieve 29.6\%, 31.1\% and 33.3\% mask AP, respectively. Our proposed one-stage methods CiCo-Yolact and CiCo-CondInst bring significant performance improvements, achieving 35.2\% and 35.4\% mask AP. Among those transformer based methods, we are unable to report the result of VisTR \cite{wang2020end} due to limited computation resources. IFC \cite{hwang2021video} obtains 36.6\% mask AP, while CiCo-Yolact and CiCo-CondInst using Swin-tiny backbone achieve the best results so far, 38.0\% and 39.1\% mask AP. 

\textbf{OVIS valid set.} Videos in OVIS are longer and have more objects per frame, resulting in more challenges, such as object occlusion, severe appearance changes, re-appeared and newly-appeared objects, and so on.
The early frame-level methods without alignment module such as MaskTrack R-CNN \cite{yang2019video} and SipMask \cite{cao2020sipmask} get very low performance (around 10.8\% mask AP), which can be 
improved by about 4~5\% mask AP by adding offline inter-frame calibration (CMaskTrack and CSipMask \cite{qi2021occluded}). STMask \cite{Li_2021_CVPR} and CrossVIS \cite{yang2021crossover} directly plug in an online inter-frame calibration, achieving similar performance. For clip-level methods, STEm-Seg \cite{Athar_Mahadevan20ECCV} obtains very low performance. In comparison, our proposed CiCo-Yolact and CiCo-CondInst achieve 17.2\% and 18.0\% mask AP, improving more than 3\% mask AP over their corresponding frame-level approaches, such as STMask \cite{Li_2021_CVPR} and CrossVIS \cite{yang2021crossover}.
Limited by our computational resources, we used four V100 GPUs to train VisTR \cite{wang2020end} with 18-frame clips on OVIS, achieving 10.2\% mask AP with 18.1 FPS. In contrast, our CiCo-Yolact and CiCo-CondInst can achieve 18.0\% and 18.2\% mask AP, recording the best result so far. 

\subsection{Results with ResNet101 Backbone}
\begin{table*}[t]
\begin{center}
\setlength{\tabcolsep}{0.5mm}{
      \linespread{2}
      \begin{tabular}
      {p{0.1\textwidth}<{\centering}p{0.12\textwidth}<{\centering}p{0.185\textwidth}p{0.04\textwidth}<{\centering}p{0.075\textwidth}<{\centering}
      p{0.055\textwidth}<{\centering}p{0.065\textwidth}<{\centering}p{0.065\textwidth}<{\centering}p{0.055\textwidth}<{\centering}p{0.065\textwidth}<{\centering}p{0.07\textwidth}<{\centering}}
         \toprule
         \ Data\ &\ Type\ & Methods &\ $T$\ &Epoch&\ AP\  &AP$_{50}$&AP$_{75}$& AR$_{1}$ & AR$_{10}$ & FPS\\
         \midrule
         \multirow{10}{*}{\shortstack{YT19}} & \multirow{7}{*}{\shortstack{Frame-\\ level}}
         &MaskProp*\cite{bertasius2020classifying}  & 13 &20   & 42.5 &  -   & 45.6  &-    &- &-\\
         &&Pro-Red*\cite{lin2021video}        & 36 &4+5  & 43.8 & 65.5 & 47.4 & 43.0 & 53.2 &- \\ 
         &&SipMask\cite{cao2020sipmask}               & 2  &12   & 35.0 & 56.1 & 35.2 & 36.0 &41.2 & 30.0\\
         &&Sg-net \cite{liu2021sg}                    & 2  &-    & 36.3 & 57.1 & 39.6 &35.9 & 43.0 &24.2\\     
         &&STMask \cite{Li_2021_CVPR}                 & 2  &12   & 36.8 & 56.8 & 38.0 & 34.8 &41.8 & 23.4\\
         &&CrossVIS \cite{yang2021crossover}          & 2  &12   & 36.6 & 57.3 & 39.7 & 36.0 &42.0& 35.6\\
         &&PCAN\cite{pcan}                            & -  &12   & 37.6 & 57.2 & 41.3 & 37.2 & 43.9 &12.7\\
         \cline{2-11}
         &\multirow{4}{*}{\shortstack{Clip- \\ level}} 
         &STEm-Seg \cite{Athar_Mahadevan20ECCV}       & 8   &-    & 34.6 & 55.8 & 37.9  &34.4 &41.6 &$<3$\\
         &&VisTR \cite{wang2020end}                   & 36  &18   & 38.6 & \textbf{61.3} & 42.3 & 37.6 &44.2 &27.7\\
         &&CiCo-Yolact                                & 3   &12   & \textbf{39.6} & 59.5 & \textbf{44.8} & \textbf{38.2} &\textbf{45.9} & 18.3\\
         &&CiCo-CondInst                              & 3   &12   & \textbf{39.6} & 60.0 & 43.7 & 37.9 & 45.6 & 18.4\\
         \midrule
         \multirow{6}{*}{\shortstack{YT21}} & \multirow{3}{*}{\shortstack{Frame-\\ level}}
         &SipMask \cite{cao2020sipmask}             & 2  &12 & 33.0 & 53.8 & 34.6 & 29.5 & 38.3 & 30.0\\
         &&STMask \cite{Li_2021_CVPR}               & 2  &12 & 34.6 & 54.0 & 38.0 & 29.4 & 39.1 &23.4\\
         &&CrossVIS \cite{yang2021crossover}        & 2  &12 & 34.6 & 54.7 & 39.6 & 30.8 & 40.8 & 35.6 \\
         \cline{2-11}
         &\multirow{3}{*}{\shortstack{Clip- \\ level}} 
         &STEm-Seg \cite{Athar_Mahadevan20ECCV}       & 3  &-   &31.1 & 52.2 & 33.1 & 29.0 & 38.3 &$<3$\\
         &&CiCo-Yolact                                & 3  &12   &36.5 & 56.1 & \textbf{41.5} & 31.6 & 41.5 &18.4\\
         &&CiCo-CondInst                              & 3  &12   &\textbf{36.7} & \textbf{56.9} & 41.2 & \textbf{32.9} & \textbf{43.0} & 18.3\\
         \midrule
         \multirow{6}{*}{\shortstack{OVIS}} & \multirow{2}{*}{\shortstack{Frame-\\ level}}
         &SipMask \cite{cao2020sipmask}             & 2  &12 &11.6 & 27.0 & 9.6 & 8.1 & 16.3 & 30.0\\
         &&STMask \cite{Li_2021_CVPR}               & 2  &12 &17.3 & 35.4 & 15.2 & 8.4 & 23.1 & 23.4\\
        %  &&CrossVIS \dag\cite{yang2021crossover}    & 2  &12 &12.9\\
         \cline{2-11}
         &\multirow{3}{*}{\shortstack{Clip- \\ level}} 
         &STEm-Seg \cite{Athar_Mahadevan20ECCV}       & 8  &-    & \textbf{20.6} & \textbf{41.7} & 18.0 & \textbf{12.3} &\textbf{27.8} &$<3$\\
         &&CiCo-Yolact                                & 3  &12   &19.1 & 37.7 & \textbf{18.7}& 10.7& 25.0& 17.1\\
         &&CiCo-CondInst                              & 3  &12   &20.4 & 41.0 & 18.1 & 11.4 & 26.4 & 17.2\\
        \bottomrule
      \end{tabular}
}
\end{center}
\vspace{-2mm}
\caption{Quantitative performance comparison of state-of-the-art methods with ResNet101 backbone on VIS benchmark datasets, where * means that the authors used their own training settings (mainly for competitions).
% and \dag means that the results with ResNet101 backbone are trained by us based on their published model settings and code. 
}\label{tab:sota_VIS_r101}
\vspace{-5mm}
\end{table*}

Table \ref{tab:sota_VIS_r101} compares the quantitative performance of state-of-the-art methods with ResNet101 backbone on all VIS benchmark datasets. Similar observations to those with ResNet50 backbone can be made.

\textbf{YouTube-VIS 2019 valid set.} The two-stage frame-level methods MaskProp \cite{bertasius2020classifying} and Proposal-Reduce \cite{lin2021video} achieve 42.5\% and 43.8\% mask AP by introducing short-term or long-term inter-frame alignment. It should be noted that MaskProp and Propose-Reduce use their own training settings (mainly for competitions) and combine multiple networks, including hybrid task cascade network and high-resolution mask refinement. The one-stage frame-level methods SipMask \cite{cao2020sipmask}, STMask \cite{Li_2021_CVPR}, Sg-net \cite{liu2021sg}, CrossVIS \cite{yang2021crossover} and PCAN \cite{pcan} obtain 35.0\%, 36.3\%, 36.8\%, 36.6\% and 37.6\% mask AP, respectively. For clip-level methods, STEm-Seg \cite{Athar_Mahadevan20ECCV} only achieves 34.6\% mask AP, and VisTR \cite{wang2020end} with the help of memory-consuming spatio-temporal global attention blocks can obtain 38.6\% mask AP, respectively. Our proposed one-stage CiCo-Yolact and Cico-CondInst with 3-frame clips as input can reach 39.6\% mask AP, achieving an improvement of 3\% mask AP compared with one-stage frame-level methods STMask \cite{Li_2021_CVPR} and CrossVIS \cite{yang2021crossover}. This demonstrates the powerful capability of CiCo to effectively exploit the temporal coherency of short video clips.   

\textbf{YouTube-VIS 2021 valid set.} These one-stage frame-in frame-out methods SipMask \cite{cao2020sipmask}, STMask \cite{Li_2021_CVPR} and CrossVIS \cite{yang2021crossover} obtain 33.0\%, 34.6\% and 34.6\% mask AP, while our proposed one-stage clip-in clip-out methods CiCo-Yolact and CiCo-CondInst can bring 2\% mask AP improvement, reaching 36.5\% and 36.7\% mask AP, respectively. However, the clip-level approach STEm-Seg \cite{Athar_Mahadevan20ECCV} only achieves 31.1\% mask AP due to its bottom-up detection paradigm.

\textbf{OVIS valid set.} This dataset focuses on instance segmentation for challenging videos with occluded objects and longer duration. 
For these early one-stage frame-level methods, SipMask \cite{cao2020sipmask} without any temporal information aggregation only obtains 11.6\% mask AP, and STMask \cite{Li_2021_CVPR} achieves 17.3\% mask AP by introducing a temporal fusion module between two adjacent frames. For clip-level methods, STEm-Seg \cite{Athar_Mahadevan20ECCV}, which is based on spatio-temporal embeddings, can keep robust and accurate tracking on longer videos and achieve 20.6\% mask AP. However, it needs to cluster pixels belonging to a specific object instance over an entire video clip, resulting in very slow speeds (below 3 FPS). In comparison, our proposed CiCo-yolact and CiCo-CondInst keep a good balance between speed and accuracy, obtaining 18.7\% and 18.2\% mask AP with around 17 FPS.

\subsection{Visual Results with ResNet50 Backbone}
\begin{figure*}[!t]
\begin{center}
    \includegraphics[width=1\linewidth]{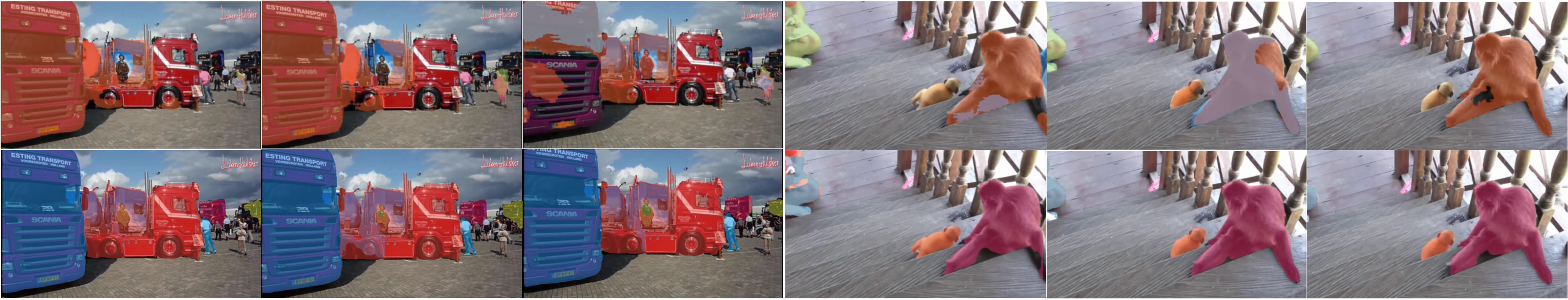}
\end{center}
\vspace{-6mm}
\caption{Visualization of instance segmentation results generated by VisTR (first row) and CiCo-Yolact (second row) on two videos of YouTube-VIS 2019 valid set. 
}\label{fig:vistr_cico}
\vspace{-2mm}
\end{figure*}
\begin{figure*}[!t]
\begin{center}
    \includegraphics[width=1\linewidth]{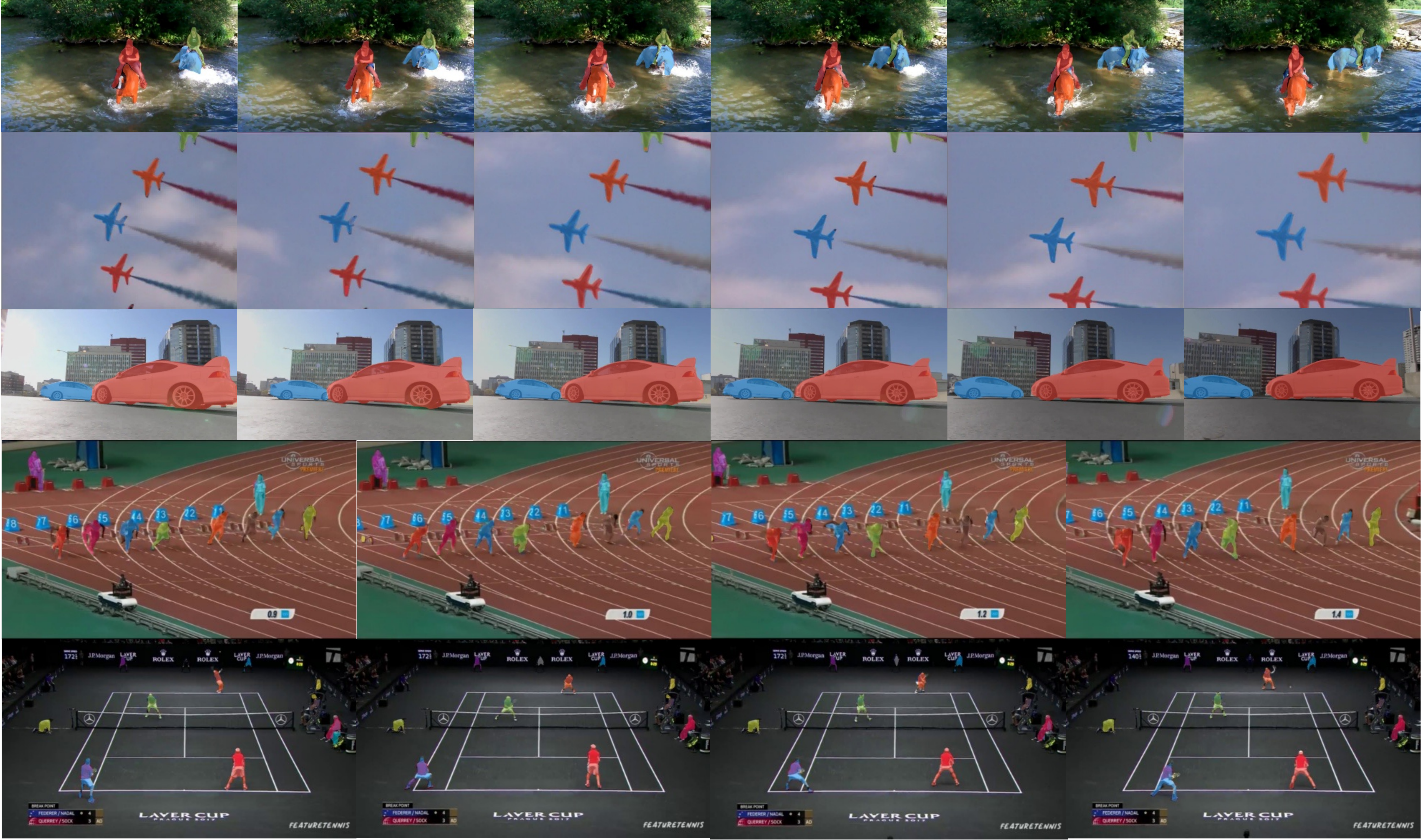}
    \includegraphics[width=1\linewidth]{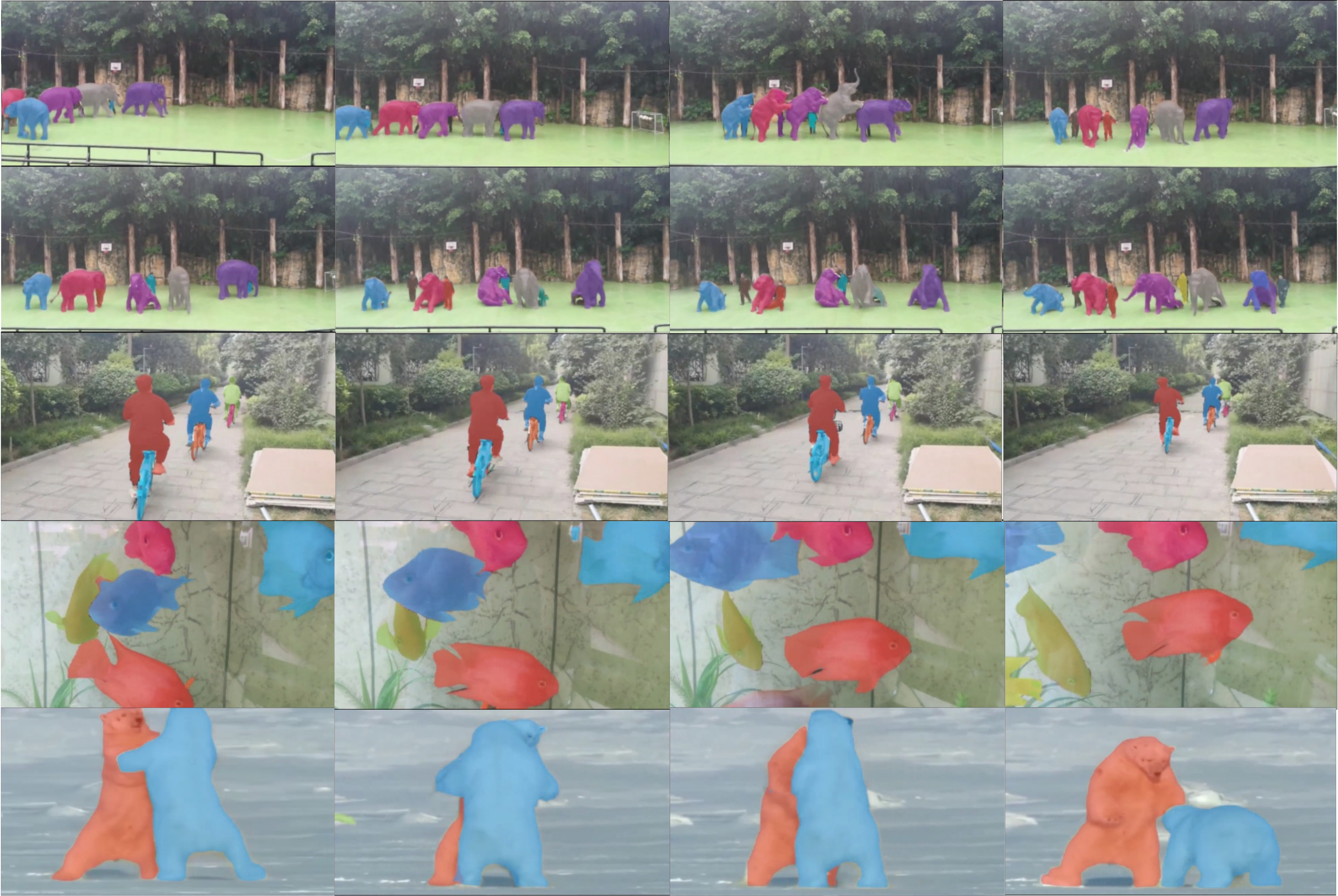}
\end{center}
\vspace{-6mm}
\caption{Visualization on instance segmentation results by CiCo-Yolact on challenging videos of YouTube-VIS 2021 (top five rows) and OVIS (bottom five rows) valid sets.}\label{fig:visual_sm}
\end{figure*}
\begin{figure*}[!t]
\begin{center}
    \includegraphics[width=1\linewidth]{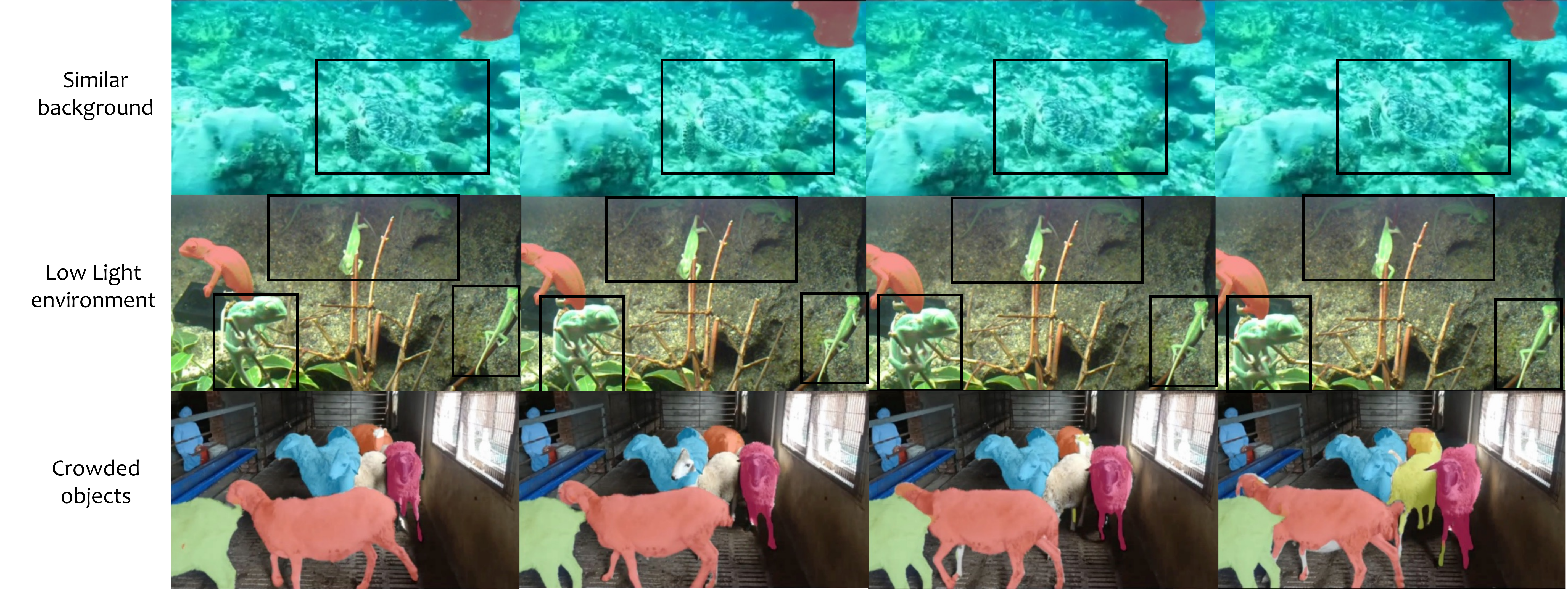}
    \includegraphics[width=1\linewidth]{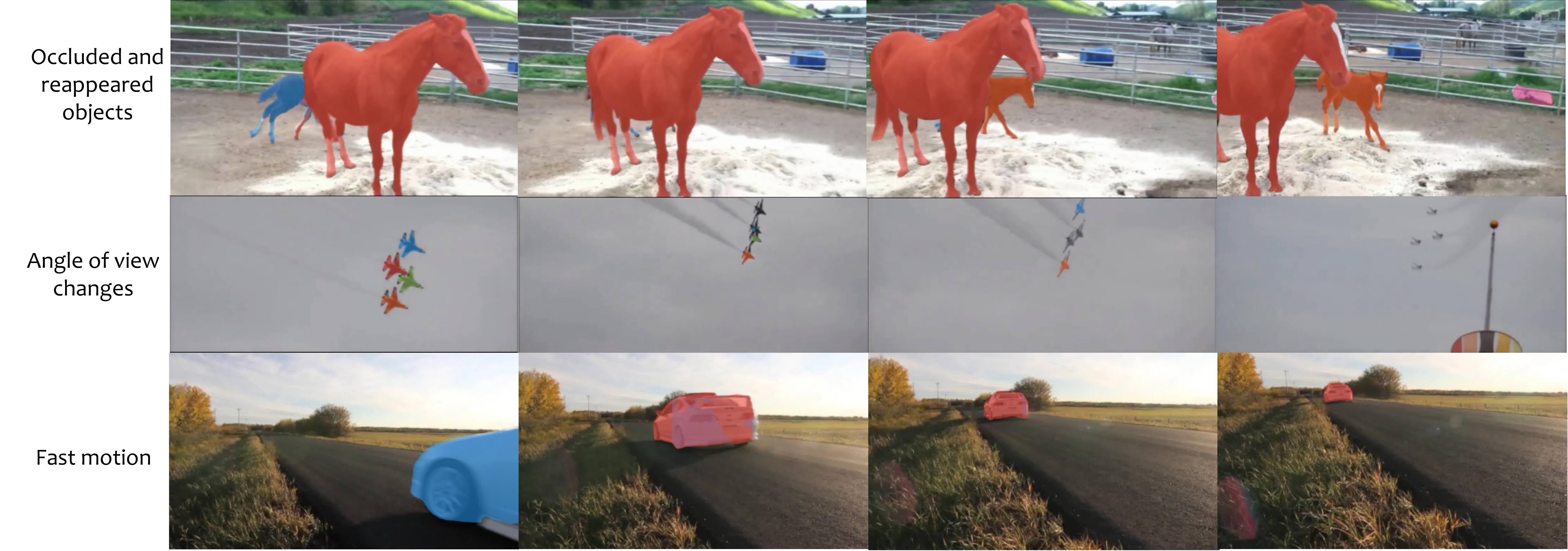}
\end{center}
\vspace{-6mm}
\caption{Failure cases. In the first two rows, the objects are missed, while in the last three rows, the objects are not well segmented mainly due to the simple cross-clip tracking strategy used in CiCo.}\label{fig:failure_cases}
\vspace{-2mm}
\end{figure*}

Fig. \ref{fig:vistr_cico} compares the instance masks produced by VisTR \cite{wang2020end} and CiCo-Yolact. One can see that CiCo can more accurately detect and segment objects in those crowded scenes. Compared with image instance segmentation, video instance segmentation encounters more challenging issues, such as uncommon camera-to-object view, motion blur, occlusion, out of focus, similar texture between objects and background, {\it etc}. Our proposed CiCo can handle some of these challenging cases well, as shown in Fig. \ref{fig:visual_sm}, including uncommon camera-to-object view, motion blur, occlusion, out of focus and small objects. Nonetheless, there are still some tricky cases where objects can not be well detected, segmented and tracked, as shown in Fig. \ref{fig:failure_cases}, including similar textures between objects and background, low light environment, tracking with fast motion. Actually, these difficult scenarios cannot be well handled by most of the existing VIS methods, and we will further investigate these issues by improving our CiCo scheme.

\vspace{-1mm}
\section{Conclusions}
\vspace{-1mm}

Based on the fact that there are high temporal coherence in adjacent frames of videos, we proposed a general strategy to extend existing one-stage frame-in frame-out (FiFo) instance segmentation approaches to clip-in clip-out (CiCo) VIS approaches. To enable CiCo performing VIS clip by clip, we expanded clip-level prediction heads and clip-level mask head by 3D convolutions to extract spatio-temporal features in a short video clip. Then, the clip-level prototypes and corresponding mask parameters were input into a small fully convolution network with dynamic filters to obtain the desired clip-level instance masks. A clip-level instance segmentation loss was proposed to train a CiCo VIS model. Two CiCo VIS models, CiCo-Yolact and CiCo-CodInst, were instantiated, and they recorded new state-of-the-art results with competitive efficiency on benchmark VIS datasets, validating the effectiveness of our CiCo approach. 
%In particular, it achieved 37.1\%, 35.3\% and 18.0\% mask AP using ResNet50 backbone, and 41.8\%, 38.9\% and xxx\% mask AP using SwinT-tiny backbone on YouTube-VIS 2019, 2021 and OVIS valid sets.

% \clearpage
% ---- Bibliography ----
\bibliographystyle{ieee_fullname}
\bibliography{main}

\begin{thebibliography}{10}\itemsep=-1pt

\bibitem{Athar_Mahadevan20ECCV}
Ali Athar, Sabarinath Mahadevan, Aljo{\v{s}}a O{\v{s}}ep, Laura Leal-Taix{\'e},
  and Bastian Leibe.
\newblock Stem-seg: Spatio-temporal embeddings for instance segmentation in
  videos.
\newblock In {\em ECCV}, 2020.

\bibitem{bertasius2020classifying}
Gedas Bertasius and Lorenzo Torresani.
\newblock Classifying, segmenting, and tracking object instances in video with
  mask propagation.
\newblock In {\em CVPR}, pages 9739--9748, 2020.

\bibitem{bolya2019yolact}
Daniel Bolya, Chong Zhou, Fanyi Xiao, and Yong~Jae Lee.
\newblock Yolact: Real-time instance segmentation.
\newblock In {\em ICCV}, pages 9157--9166, 2019.

\bibitem{cao2020sipmask}
Jiale Cao, Rao~Muhammad Anwer, Hisham Cholakkal, Fahad~Shahbaz Khan, Yanwei
  Pang, and Ling Shao.
\newblock {SipMask}: Spatial information preservation for fast image and video
  instance segmentation.
\newblock {\em arXiv preprint arXiv:2007.14772}, 2020.

\bibitem{carion2020end}
Nicolas Carion, Francisco Massa, Gabriel Synnaeve, Nicolas Usunier, Alexander
  Kirillov, and Sergey Zagoruyko.
\newblock End-to-end object detection with transformers.
\newblock In {\em ECCV}, pages 213--229. Springer, 2020.

\bibitem{chen2020blendmask}
Hao Chen, Kunyang Sun, Zhi Tian, Chunhua Shen, Yongming Huang, and Youliang
  Yan.
\newblock {BlendMask}: Top-down meets bottom-up for instance segmentation.
\newblock In {\em CVPR}, pages 8573--8581, 2020.

\bibitem{chen2018masklab}
Liang-Chieh Chen, Alexander Hermans, George Papandreou, Florian Schroff, Peng
  Wang, and Hartwig Adam.
\newblock {MaskLab}: Instance segmentation by refining object detection with
  semantic and direction features.
\newblock In {\em CVPR}, pages 4013--4022, 2018.

\bibitem{dai2016instance}
Jifeng Dai, Kaiming He, and Jian Sun.
\newblock Instance-aware semantic segmentation via multi-task network cascades.
\newblock In {\em CVPR}, pages 3150--3158, 2016.

\bibitem{dai2017deformable}
Jifeng Dai, Haozhi Qi, Yuwen Xiong, Yi Li, Guodong Zhang, Han Hu, and Yichen
  Wei.
\newblock Deformable convolutional networks.
\newblock In {\em ICCV}, pages 764--773, 2017.

\bibitem{QueryInst}
Yuxin Fang, Shusheng Yang, Xinggang Wang, Yu Li, Chen Fang, Ying Shan, Bin
  Feng, and Wenyu Liu.
\newblock Instances as queries.
\newblock {\em arXiv preprint arXiv:2105.01928}, 2021.

\bibitem{fu2020compfeat}
Yang Fu, Linjie Yang, Ding Liu, Thomas~S Huang, and Humphrey Shi.
\newblock Compfeat: Comprehensive feature aggregation for video instance
  segmentation.
\newblock {\em arXiv preprint arXiv:2012.03400}, 6, 2020.

\bibitem{he2017mask}
Kaiming He, Georgia Gkioxari, Piotr Doll{\'a}r, and Ross Girshick.
\newblock Mask {R-CNN}.
\newblock In {\em ICCV}, pages 2961--2969, 2017.

\bibitem{he2016deep}
Kaiming He, Xiangyu Zhang, Shaoqing Ren, and Jian Sun.
\newblock Deep residual learning for image recognition.
\newblock In {\em CVPR}, pages 770--778, 2016.

\bibitem{huang2019mask}
Zhaojin Huang, Lichao Huang, Yongchao Gong, Chang Huang, and Xinggang Wang.
\newblock Mask scoring {R-CNN}.
\newblock In {\em CVPR}, pages 6409--6418, 2019.

\bibitem{hwang2021video}
Sukjun Hwang, Miran Heo, Seoung~Wug Oh, and Seon~Joo Kim.
\newblock Video instance segmentation using inter-frame communication
  transformers.
\newblock {\em arXiv preprint arXiv:2106.03299}, 2021.

\bibitem{pcan}
Lei Ke, Xia Li, Martin Danelljan, Yu-Wing Tai, Chi-Keung Tang, and Fisher Yu.
\newblock Prototypical cross-attention networks for multiple object tracking
  and segmentation.
\newblock In {\em NeurIPS}, 2021.

\bibitem{Li_2021_CVPR}
Minghan Li, Shuai Li, Lida Li, and Lei Zhang.
\newblock Spatial feature calibration and temporal fusion for effective
  one-stage video instance segmentation.
\newblock In {\em CVPR}, pages 11215--11224, 2021.

\bibitem{li2017fully}
Yi Li, Haozhi Qi, Jifeng Dai, Xiangyang Ji, and Yichen Wei.
\newblock Fully convolutional instance-aware semantic segmentation.
\newblock In {\em CVPR}, pages 2359--2367, 2017.

\bibitem{lin2021video}
Huaijia Lin, Ruizheng Wu, Shu Liu, Jiangbo Lu, and Jiaya Jia.
\newblock Video instance segmentation with a propose-reduce paradigm.
\newblock {\em arXiv preprint arXiv:2103.13746}, 2021.

\bibitem{lin2017feature}
Tsung-Yi Lin, Piotr Doll{\'a}r, Ross Girshick, Kaiming He, Bharath Hariharan,
  and Serge Belongie.
\newblock Feature pyramid networks for object detection.
\newblock In {\em CVPR}, pages 2117--2125, 2017.

\bibitem{lin2014microsoft}
Tsung-Yi Lin, Michael Maire, Serge Belongie, James Hays, Pietro Perona, Deva
  Ramanan, Piotr Doll{\'a}r, and C~Lawrence Zitnick.
\newblock Microsoft {COCO}: Common objects in context.
\newblock In {\em ECCV}, pages 740--755. Springer, 2014.

\bibitem{liu2021sg}
Dongfang Liu, Yiming Cui, Wenbo Tan, and Yingjie Chen.
\newblock Sg-net: Spatial granularity network for one-stage video instance
  segmentation.
\newblock In {\em CVPR}, pages 9816--9825, 2021.

\bibitem{liu2018path}
Shu Liu, Lu Qi, Haifang Qin, Jianping Shi, and Jiaya Jia.
\newblock Path aggregation network for instance segmentation.
\newblock In {\em CVPR}, pages 8759--8768, 2018.

\bibitem{Liu_2021_ICCV}
Ze Liu, Yutong Lin, Yue Cao, Han Hu, Yixuan Wei, Zheng Zhang, Stephen Lin, and
  Baining Guo.
\newblock Swin transformer: Hierarchical vision transformer using shifted
  windows.
\newblock In {\em ICCV}, pages 10012--10022, October 2021.

\bibitem{liu2021Swin}
Ze Liu, Yutong Lin, Yue Cao, Han Hu, Yixuan Wei, Zheng Zhang, Stephen Lin, and
  Baining Guo.
\newblock Swin transformer: Hierarchical vision transformer using shifted
  windows.
\newblock In {\em ICCV}, 2021.

\bibitem{qi2021occluded}
Jiyang Qi, Yan Gao, Yao Hu, Xinggang Wang, Xiaoyu Liu, Xiang Bai, Serge
  Belongie, Alan Yuille, Philip Torr, and Song Bai.
\newblock Occluded video instance segmentation: Dataset and challenge.
\newblock In {\em NeurIPS}, 2021.

\bibitem{sun2021sparse}
Peize Sun, Rufeng Zhang, Yi Jiang, Tao Kong, Chenfeng Xu, Wei Zhan, Masayoshi
  Tomizuka, Lei Li, Zehuan Yuan, Changhu Wang, et~al.
\newblock Sparse r-cnn: End-to-end object detection with learnable proposals.
\newblock In {\em CVPR}, pages 14454--14463, 2021.

\bibitem{tian2020conditional}
Zhi Tian, Chunhua Shen, and Hao Chen.
\newblock Conditional convolutions for instance segmentation.
\newblock {\em arXiv preprint arXiv:2003.05664}, 2020.

\bibitem{vaswani2017attention}
Ashish Vaswani, Noam Shazeer, Niki Parmar, Jakob Uszkoreit, Llion Jones,
  Aidan~N Gomez, {\L}ukasz Kaiser, and Illia Polosukhin.
\newblock Attention is all you need.
\newblock In {\em NeurIPS}, pages 5998--6008, 2017.

\bibitem{wang2021end}
Tao Wang, Ning Xu, Kean Chen, and Weiyao Lin.
\newblock End-to-end video instance segmentation via spatial-temporal graph
  neural networks.
\newblock In {\em ICCV}, pages 10797--10806, 2021.

\bibitem{wang2020end}
Yuqing Wang, Zhaoliang Xu, Xinlong Wang, Chunhua Shen, Baoshan Cheng, Hao Shen,
  and Huaxia Xia.
\newblock End-to-end video instance segmentation with transformers.
\newblock In {\em CVPR}, 2021.

\bibitem{yang2019video}
Linjie Yang, Yuchen Fan, and Ning Xu.
\newblock Video instance segmentation.
\newblock In {\em ICCV}, pages 5188--5197, 2019.

\bibitem{yang2021crossover}
Shusheng Yang, Yuxin Fang, Xinggang Wang, Yu Li, Chen Fang, Ying Shan, Bin
  Feng, and Wenyu Liu.
\newblock Crossover learning for fast online video instance segmentation.
\newblock In {\em ICCV}, pages 8043--8052, 2021.

\end{thebibliography}
\end{document}